\documentclass[10pt,twocolumn,letterpaper]{article}

\usepackage{titletoc}

\usepackage{wacv}              %

\usepackage{graphicx}
\usepackage{amsmath}
\usepackage{amssymb}
\usepackage{booktabs}
\usepackage{colortbl}
\usepackage{xcolor}
\usepackage{multirow}
\usepackage{makecell}
\usepackage{stmaryrd}
\usepackage{mathtools}

\usepackage[pagebackref,breaklinks,colorlinks]{hyperref}

\usepackage[all=normal, mathspacing=normal,mathdisplays=normal, floats=tight, paragraphs=tight, lists=normal]{savetrees}

\usepackage[capitalize]{cleveref}
\crefname{section}{Sec.}{Secs.}
\Crefname{section}{Section}{Sections}
\Crefname{table}{Table}{Tables}
\crefname{table}{Tab.}{Tabs.}

\newcommand{\second}{\cellcolor{blue!10}}
\newcommand{\first}{\cellcolor{blue!30}}
\newcommand{\tablesep}{\cellcolor[rgb]{0.900,0.900,0.900}}
\definecolor{cvprblue}{rgb}{0.21,0.49,0.74}

\def\wacvPaperID{867} %
\def\confName{WACV}
\def\confYear{2025}

\begin{document}

\title{Hierarchical Light Transformer Ensembles\\ for Multimodal Trajectory Forecasting}

\author{\textbf{Adrien Lafage,\textsuperscript{\rm 1,2,$\dagger$} Mathieux Barbier,\textsuperscript{\rm 2} Gianni Franchi\textsuperscript{\rm 1} \& David Filliat\textsuperscript{\rm 1}}\\
U2IS, ENSTA Paris, Institut Polytechnique de Paris\textsuperscript{\rm 1}\\
Ampere Software Technology\textsuperscript{\rm 2}
}

\maketitle

\def\thefootnote{$\dagger$}
\begin{NoHyper}
\footnotetext{{\tt adrien.lafage@ensta-paris.fr} \hfill}
\end{NoHyper}
\def\thefootnote{\arabic{footnote}}

\begin{abstract}
   Accurate trajectory forecasting is crucial for the performance of various systems, such as advanced driver-assistance systems and self-driving vehicles. These forecasts allow us to anticipate events that lead to collisions and, therefore, to mitigate them. Deep Neural Networks have excelled in motion forecasting, but overconfidence and weak uncertainty quantification persist. Deep Ensembles address these concerns, yet applying them to multimodal distributions remains challenging.
   In this paper, we propose a novel approach named Hierarchical Light Transformer Ensembles (HLT-Ens) aimed at efficiently training an ensemble of Transformer architectures using a novel hierarchical loss function. HLT-Ens leverages grouped fully connected layers, inspired by grouped convolution techniques, to capture multimodal distributions effectively. We demonstrate that HLT-Ens achieves state-of-the-art performance levels through extensive experimentation, offering a promising avenue for improving trajectory forecasting techniques. We make our code available at \href{https://github.com/alafage/hlt-ens}{github.com/alafage/hlt-ens}.
\end{abstract}

\section{Introduction}
\label{sec:intro}

\begin{figure}[t]
  \centering
   \includegraphics[width=\linewidth]{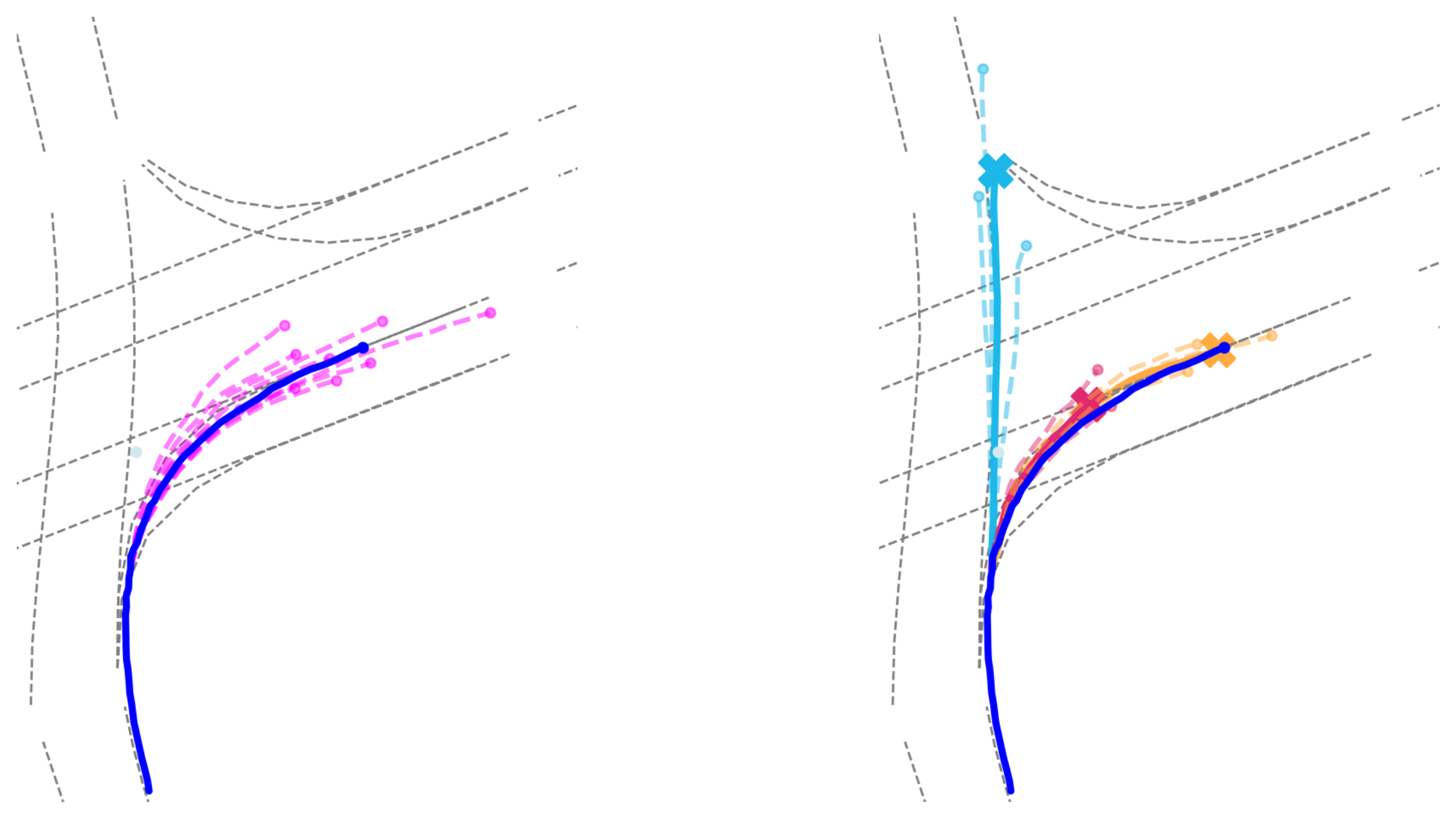}

   \caption{\textbf{Illustration of a hierarchical mixture (\textit{right}) compared to a classical one (\textit{left}) on a sample from the Argoverse 1 dataset}. Ground truth is in dark blue in both figures, forecasts are colored dashed lines, and grey dashed lines are the centerlines of road lanes. On the right figure, we display the \textit{meta-modes} (3 solid lines ending with a cross) inferred from the predictions and their associated predictions (3 of the same color). The hierarchical structure notably enables efficient prediction compression by taking only the \textit{meta-modes}.}
   \label{fig:tease}
\end{figure}

Trajectory forecasting is a cornerstone of Advanced Driver-Assistance Systems (ADAS) and Autonomous Vehicles (AV), enabling them to anticipate effectively future risks and make informed decisions \cite{cui2019multimodal,bengler2014three}. However, accurately predicting trajectories is challenging due to the multimodal nature of the problem, where drivers can exhibit diverse behaviors in various situations (e.g., at an intersection).
Many existing methods, deployed in real-world systems, achieve satisfactory performance primarily for short-term prediction horizons, typically employing kinematics models \cite{lefevre2014survey}. Yet, these models struggle to incorporate contextual information and adapt to sudden variations in traffic agent motion.

Indeed, achieving a longer prediction time horizon is challenging as the model has to account for complex interactions between varying numbers of traffic agents. As the uncertainty increases when forecasting further in time, predicting a single development becomes less relevant \cite{lee2017desire,deo2018convolutional,gupta2018social_gan}. In this context, providing multiple outcomes (i.e., a set of forecasts) is more attractive but adds complexity to the task \cite{gupta2018social_gan, girgis2022latent,varadarajan2022multipath++}. A prevalent approach involves the utilization of Mixture Density Networks (MDNs) \cite{bishop1994mixture} to capture the diverse modes (i.e., local maxima) of trajectory distributions efficiently \cite{girgis2022latent,ngiam2021scene,nayakanti2023wayformer,zhou2023query}. However, such methods face challenges in determining the appropriate number of modes, often ill-defined, and ensuring that all the identified modes are distinct. For instance, when approaching an intersection (see \cref{fig:tease}), there could be two modes corresponding to turning right or proceeding straight. But, additional \textit{sub-modes} may arise based on driver experiences or contextual factors.
Furthermore, their industrial applications to safety-critical systems still need to be improved as they exhibit several shortcomings, such as overconfidence, which is often typical in Deep Neural Networks (DNNs) \cite{nguyen2015deep,guo2017calibration,mcallister2017concrete}. A research direction to mitigate these issues is ensembling. Ensembles usually enable the improvement of the model's accuracy \cite{izmailov2018averaging} and strengthen the resilience to uncertainty \cite{lakshminarayanan2017simple}. Although they have been successfully applied to trajectory forecasting tasks \cite{varadarajan2022multipath++,wang2022tenet,girgis2022latent,zhou2023query}, their advantages come with increased computational overhead at both the training and evaluation time.

We propose a novel solution to address these challenges based on hierarchical multi-modal density networks (\cref{fig:tease}, right) and efficient ensemble techniques. Our approach introduces a new loss function that presents a hierarchical structure in multi-modal predictions, allowing us to represent traffic agents' diverse behaviors better. Additionally, based on previous work \cite{laurent2023packed}, we adapt the concept of Packed-Ensembles, initially designed for Convolutional Neural Networks, to the realm of transformer-based models and develop a framework to train ensembles of Transformers efficiently without increasing computational complexity. 

We demonstrate the effectiveness of our approach through extensive experimentation on trajectory forecasting datasets \cite{chang2019argoverse,zhan2019interaction}, showcasing improved performance compared to existing methods. In summary, our contributions are the following:
\begin{itemize}
    \item We introduce a novel density representation and associated loss function tailored for optimizing MDNs, leveraging a hierarchical structure to capture the nuanced nature of modes in trajectory forecasting and to embed spatial information in the prediction.
    \item We devise a light ensemble framework tailored explicitly for Transformer-based architectures, significantly reducing computational overhead while maintaining forecast quality.
    \item Our approach, \textit{Hierarchical Light Transformer Ensembles} (HLT-Ens), offers versatility by seamlessly adapting to diverse architectures, enabling them to improve their performance. 
\end{itemize}

\section{Related Works}\label{sec:related_works}

\subsection{Multimodal Trajectory Forecasting}
Predicting a single forecast or unimodal distribution is insufficient when producing long-term forecasting. %
The future position distribution will likely follow a multimodal distribution \cite{deo2018convolutional,gupta2018social_gan,lee2017desire,girgis2022latent,zhou2023query}.
For instance, without any prior information, the distribution density of vehicle future positions should be higher on the roads and lower outside. At the entry of an intersection, it might be relevant to estimate a trajectory for each possible direction. To that end, \cite{lefevre2014survey,schreier2016integrated} approximate the modes of the distribution using several unimodal models, and \cite{tang2019multiple,girgis2022latent,bahari2021injecting,cui2019multimodal,zhou2023query}, defines latent variables conditioning the decoder to generate a set of forecasts.

Yet, the predicted modes do not provide any information about the uncertainty around them. Thus, researchers have used probabilistic approaches to better approximate aleatoric uncertainty (input data-dependent errors). In particular, generative models have been widely used. Leveraging \textit{implicit} models, \cite{lee2017desire,ma2019wasserstein,salzmann2020trajectronplusplus,yuan2021agent,mangalam2020journey} define a sampling procedure to generate trajectories as if they were coming from the actual conditional distribution. However, there is no information on the number of samples required to cover the underlying distribution, limiting the application of such models on safety-critical applications.
On the other hand, \cite{varadarajan2022multipath++,chai2020multipath,deo2018convolutional,deo2018multi_modal,mercat2020multi_head,ivanovic2019trajectron} estimate explicitly the distribution density with Mixture Density Networks (MDNs) \cite{bishop1994mixture}. Although they enable flexibility to approximate uncertainties, such models are difficult to train as they suffer from training instabilities and mode collapse (i.e., several component probabilities tend to zero consistently across examples), for which we propose solutions.

On another note, there has been a growing interest in leveraging attention mechanisms to model interactions between agents and the environment, resulting in more Transformer-based approaches among the most efficient models \cite{ngiam2021scene,girgis2022latent,nayakanti2023wayformer,aydemir2023adapt,zhou2022hivt} on which we focus as our approach can be applied to any transformer backbone.

\subsection{Multiple Choice Learning}

In multimodal trajectory forecasting, we are interested in learning several acceptable hypotheses rather than one. Introduced by \cite{guzman2012multiple}, Multiple Choice Learning is a general framework for this task. It offers a means to create cooperation between multiple predictor heads. Eventually, each head becomes specialized on a subset of the data. Inspired by this approach, \cite{lee2015m} adapted it to train actual and implicit ensembles of DNNs. In particular, they introduced the \textit{Stochastic Multiple Choice Learning} (sMCL) \cite{lee2016stochastic} training scheme based on the Winner-Takes-All (WTA) loss. Although the dataset has only one realization for each input, it enables diversity among the heads. With the same objective, \cite{rupprecht2017learning} use a shared architecture instead of several independent ones and demonstrate the benefits of MCL for multi-label tasks while having significantly fewer parameters than its ensemble counterpart. Interestingly, many recent works leveraging MDNs for multimodal trajectory forecasting employ WTA loss variants, easing the mixture optimization \cite{zhou2023query,wang2022tenet,makansi2019overcoming}.
Ideally, the WTA loss leads to a Voronoi tesselation of the ground truth \cite{rupprecht2017learning}, which could be interpreted as a perfect k-means clustering. %
However, similarly to the k-means algorithm, its performance depends on the hypotheses' initialization.
Indeed, the WTA loss is known to suffer from stability issues during training, referred to as hypothesis collapse, i.e., some hypotheses receive little to no gradients due to unfortunate initialization and better other hypotheses.
To mitigate this issue, \cite{rupprecht2017learning} relaxes the WTA formulation ($\varepsilon$-WTA) to ensure all hypotheses are slightly updated unconditionally to their performance. Yet, \cite{makansi2019overcoming} showcase it results in non-winning hypotheses being slowly pushed to the equilibrium, creating a spurious mode in the distribution. Therefore, they introduce the Evolving WTA (EWTA) strategy, updating the top-$n$ winners instead of solely the best one and decreasing $n$ iteratively during training. To better estimate multimodal distributions, we introduce a hierarchy to capture diversity on various scales and help the optimization process.

\subsection{Efficient Ensembling}

As previously stipulated, ensembling techniques, such as \textit{Deep Ensemble} \cite{lakshminarayanan2017simple} (DE), have appealing properties to enhance performance through the diversity of several independent networks. Still, the ensemble number of parameters linearly increases with its size. Implicit ensembles designate a group of methods attempting to mimic DE diversity while having much fewer parameters. In this line of work, BatchEnsemble \cite{Wen2019batchensemble} induces diversity in a single backbone by efficiently applying rank 1 perturbation matrices, each corresponding to a member. The multi-input multi-output (MIMO) approach \cite{havasi2021training} exploits the over-parametrization of DNNs by implicitly defining several subnetworks within one backbone through MIMO settings. Recently, \cite{laurent2023packed} have highlighted the possibility of creating ensembles of smaller networks performing comparably with DE. However, these approaches are applied to image classification tasks based on Convolutional Neural Networks (CNN). In our work, we present an efficient way to ensemble Transformer architecture.

Ensembling has been applied to trajectory forecasting problems. Yet the original formulation of Deep Ensembles for regression tasks is not suited to the Multiple Choice Learning framework where we want to retain diversity. Instead, we are interested in finding an efficient method to use sets of predictions to create a single set. In \cite{malinin2021shifts}, the authors present the Robust Imitative Planning (RIP) algorithm, which keeps the hypotheses having the higher likelihoods on the estimated distributions from the ensemble. The issue is that it heavily relies on the quality of the predicted distributions and does not ensure any diversity. For better coverage, clustering approaches such as the Expectation-Maximization \cite{varadarajan2022multipath++} or the K-Means \cite{wang2022tenet,zhou2023query,nayakanti2023wayformer} algorithms are attractive. In our experiments, we provide a benchmark of these ensembling strategies.

\section{Background}
\label{sec:background}

In this section, we present the mathematical formalism for this work and offer a brief background on training MDNs for trajectory forecasting. Appendix \textbf{A} summarizes the main notations.

\subsection{Trajectory Forecasting}

Let us consider a dataset $\mathcal{D}=\{\mathbf{X}^{i}\}^{|\mathcal{D}|}_{i=1}$, where each instance $\mathbf{X}^{i}\in \mathbb{R}^{T\times A\times 2}$ represents the trajectories of $A$ traffic agents over $T$ timesteps: $\mathbf{X}^{i}=\mathbf{X}^{i}_{1:T}=(\mathbb{X}^{i}_{1}, ..., \mathbb{X}^{i}_{T})$. Here, $\mathbb{X}^{i}_{t} = (\mathbf{x}^{i}_{1}(t), ..., \mathbf{x}^{i}_{A}(t))$ with $x^{i}_{a}(t)\in \mathbb{R}^{2}$ denotes the position in the 2-dimensional Euclidean space of agent $a \in \llbracket1, A\rrbracket$ at time $t\in \mathbb{N}$ for the $i$th instance in $\mathcal{D}$.

Given an observation time-horizon $t_{obs} \in \llbracket1,T-1\rrbracket$, we define $X_i\coloneqq\mathbf{X}^{i}_{1:t_{obs}} \in \mathbb{R}^{t_{obs}\times A\times 2}$ and $Y_i\coloneqq\mathbf{X}^{i}_{t_{obs}+1:T}\in \mathbb{R}^{t_{pred}\times A\times 2}$ with $t_{pred} = T-t_{obs}$ for better legibility. Additionally, we denote $\mathbb{X}\coloneqq\mathbb{R}^{t_{obs}\times A\times 2}$, $\mathbb{Y}\coloneqq\mathbb{R}^{t_{pred}\times A\times 2}$, and $P_{\mathcal{X}, \mathcal{Y}}$ the probability followed by the dataset samples. The trajectory forecasting problem aims to estimate $Y_i$ given $X_i$ using a parametric model $f_{\theta}: \mathbb{X} \to \mathbb{Y}$ of parameter $\theta \in \mathbb{R}^{n}$. Particularly, we seek the optimal parameters $\theta^{\star}$ that minimize the expected error:
\begin{equation}\label{eq:expected_error}
    \frac{1}{|\mathcal{D}|} \sum^{|\mathcal{D}|}_{i=1}\ell(f_{\theta}(X_i), Y_i) \ \ \text{with }\ell\text{ a loss function}
\end{equation}

For a large cardinality of $\mathcal{D}$, the minimum of \cref{eq:expected_error} represents the conditional expectation, i.e., $f_{\theta^{\star}}(X_i) = \mathbb{E}\left[\mathcal{Y}|\mathcal{X}=X_i\right]$. However, while suitable for unimodal distributions, this average may perform poorly for multimodal ones, potentially overlooking low probability density areas between modes. Given the multimodal nature of vehicle motion distributions, especially in complex scenarios such as road intersections, a multiple-mode (or modal) approach is required.

\subsection{Multiple Choice Learning}

Within the Multiple Choice Learning framework, we aim to better capture phenomena' complexity by replacing a single prediction with multiple ones. Consider a %
 predictor providing $K$ estimates:
\begin{equation}
    f_{\theta}(X_i) = \left(f_{\theta}^{1}(X_i),...,f_{\theta}^{K}(X_i)\right)
\end{equation}

A commonly used loss to optimize such models is the  \textit{Winner-Takes-All} (WTA) loss (or \textit{oracle loss}), which computes the error $\ell$ for the closest prediction among the $K\in \mathbb{N}$ alternatives. As outlined by \cite{rupprecht2017learning}, this loss is defined by:
\begin{equation}\label{eq:wta}
    \int_{\mathbb{X}}\sum^{K}_{k=1}\int_{\mathbb{Y}^k(X)} \ell\left(f_{\theta}^{k}(X), Y\right)p(X, Y)dYdX
\end{equation}

where we consider a Voronoi tesselation of the space $\mathbb{Y}=\cup^{K}_{k=1}\mathbb{Y}^{k}$ induced by $K$ unknown generators $g^{k}(X)$ and the loss $\ell$:
\begin{equation}
    \mathbb{Y}^{k}(X) = \{Y \in \mathbb{Y}\ \colon\ \ell(g^{k}(X), Y) < \ell(g^{j}(X), Y)\ \forall j\neq k \}
\end{equation}

As described in \cite{rupprecht2017learning}, \cref{eq:wta} can be understood as the aggregation of the Voronoi cell losses generated by the predictions $f_{\theta}^{k}(X)$. To help understand the behavior of this loss, \cite{rupprecht2017learning} assume $\ell$ is the $\ell_{2}-loss$ commonly used in regression problems and demonstrate \cref{eq:wta} is minimal when the generators $g^{k}(X)$ and the predictors $f_{\theta}^{k}(X)$ are identical, and correspond to a centroidal Voronoi tesselation (cf. Theorem 1 of \cite{rupprecht2017learning}). Hence, minimizing this loss can be seen as a clustering in the $\mathbb{Y}$ space conditioned on some $X\in \mathbb{X}$. Applied to trajectory forecasting, it produces diverse, suitable forecasts for the same observation.

\subsection{Mixture Density Networks}

While having a set of forecasts helps in better trajectory estimation, we can further enhance our prediction by estimating uncertainties around each hypothesis using Mixture Density Networks (MDNs) \cite{bishop1994mixture}.
Drawing inspiration from Gaussian Mixture distributions, traditional MDNs enable uncertainty quantification by representing forecasts as a mixture of $K\in\mathbb{N}$ Gaussian distributions (components) whose parameters are predicted by DNNs. However, for trajectory forecasting, Laplace Mixture distributions are often preferred \cite{zhou2023query,ngiam2021scene,zhou2022hivt}. Thus, given an input $X\in\mathbb{R}^{t_{obs}\times A\times 2}$, an MDN forecasting model estimates the parameters of:
\begin{equation}
P(\mathcal{Y}|\mathcal{X}=X) = \sum_{k=1}^{K} \pi_{\theta_k}(X) \mbox{Laplace}(\boldsymbol{\mu}_{\theta_k}(X), \mathbf{b}_{\theta_k}(X))
\end{equation}
Here, $\boldsymbol{\mu}_{\theta_k}(X)\in\mathbb{R}^{t_{pred}\times A\times 2}$ represents the mean of one mode of agents' positions at each time step, $\mathbf{b}_{\theta_k}(X)$ denotes the associated scale parameter, and $\boldsymbol{\pi}_{\theta}(X)$ the mixture weights $(\pi_{\theta_1}(X),...,\pi_{\theta_k}(X)) \in\boldsymbol{\Delta}^{K}$ with $\boldsymbol{\Delta}^{C}=\{s\in\left[0,1\right]^{C}\mid \sum^{C}_{j=1} s_{j}=1\}$ the probability simplex.

\noindent

As detailed in \cite{makansi2019overcoming}, one can train MDNs using optimization strategies coming from the Multiple Choice Learning framework by considering $f_{\theta}^{k}(X)$ to be the tuple $(\boldsymbol{\mu}_{\theta_k}(X), \mathbf{b}_{\theta_k}(X), \pi_{\theta_k}(X))$ and $\ell$ to be the Negative Log Likelihood (NLL).

\section{Method}\label{sec:method}

\subsection{Hierarchical Multimodal Distribution}

Trajectory prediction involves generating multi-modal forecasts using DNNs. However, determining the appropriate number of modes (denoted as $K$) for training poses a challenge as it can vary due to multiple factors and lacks intuitive guidance. Additionally, the inherent multi-modal and uncertain nature of DNNs often leads to models with significant variances, resulting in unreliable predictions. To address this challenge and improve prediction stability, we propose adapting the MDN framework to a Mixture of Mixtures, thereby introducing a novel architecture: Hierarchical Mixture of Density Networks (HMDN).

Hence, in our proposed approach, we establish a two-level hierarchy within the multimodal distribution by introducing the concept of \textit{meta-modes} (\cref{fig:tease}, right). A \textit{meta-mode} is constructed from classical modes and is intended to approximate them. Practically, \textit{meta-modes} represent the mode cluster centroids. The mixture formed by these meta-modes is termed the \textit{meta-mixture}, we have $K^{\star}$ components within the \textit{meta-mixture}:
\begin{equation}
\resizebox{\linewidth}{!}{
$P_{k^{\star}}(\mathcal{Y}|\mathcal{X}=X) = \sum_{k’=1}^{K’} \pi_{\theta_{k^{\star},k'}}(X) \mbox{Laplace}(\boldsymbol{\mu}_{\theta_{k^{\star},k'}}(X), \mathbf{b}_{\theta_{k^{\star},k'}}(X))$
}
\end{equation}

We note $\theta_{k^{\star},k'}$ the parameters of the $k'$-th component of the $k^{\star}$-th \textit{meta-mode}. Consequently, the hierarchical multimodal distribution is equal to :
\begin{equation}
P(\mathcal{Y}|\mathcal{X}=X) = \sum_{k^{\star}=1}^{K^{\star}} \pi_{\theta_{k^{\star}}}(X) P_{k^{\star}}(\mathcal{Y}|\mathcal{X}=X)
\end{equation}

where $\pi_{\theta_{k^{\star}}}(X) = \sum_{k'\in K'} \pi_{\theta_{k^{\star},k'}}(X)$. We assume $K'$ remains consistent across all \textit{meta-modes}, although it could vary for each. Therefore, the total number of modes is $K=K^{\star}\times K'$. Furthermore, we define the mean and scale of the meta-modes respectively as $\overline{\boldsymbol{\mu}}_{\theta_{k^{\star}}}(X) = \frac{1}{K'}\sum_{k'} \pi_{\theta_{k^{\star},k'}}(X)\boldsymbol{\mu}_{\theta_{k^{\star},k'}}(X)$ and $ \overline{\mathbf{b}}_{\theta_k^{\star}}^2(X) = \frac{1}{2K'}\sum_{k'} \left(2\mathbf{b}_{\theta_{k^{\star},k'}}^{2}(X) + \right.$ 
$  \left.  \boldsymbol{\mu}_{\theta_{k^{\star},k'}}^{2}(X)\right) -  \overline{\boldsymbol{\mu}}_{\theta_{k^{\star}}}^2(X)$. Introducing these meta-modes leads to more stable predictions, as shown later in \cref{tab:stdev}. We analyze further the loss robustness to the variation of the mode number in Appendix \textbf{F}. This hierarchical structure enables more spatial information as \textit{sub-modes} can be expected to be close when sharing the same \textit{meta-mode}. The following section describes the loss we developed to train an MDN with the concept of \textit{meta-modes}.

\subsection{Hierarchical Multimode Loss}

Many existing MDN architectures rely on techniques similar to clustering, which can introduce challenges regarding robustness to changes in mode number. In contrast, our proposed approach embraces a hierarchical structure, enabling more flexibility.
To integrate this hierarchical structure into our prediction mixture, we introduce a novel loss function aimed at softly enforcing proximity among modes belonging to the same \textit{meta-mode}, termed the \textit{Hierarchical Winner-Takes-All} (\textbf{HWTA}) loss. The \textbf{HWTA} loss, denoted as $\mathcal{L}$, comprises two distinct components: the meta-mode loss ($\mathcal{L}_{\text{meta}}$) and the Meta-Winner-Takes-All (MWTA) loss ($\mathcal{L_{\text{MWTA}}}$).
\begin{equation}\label{eq:hwta}
    \mathcal{L} = \gamma \times \mathcal{L_{\text{meta}}} + (1-\gamma)\times\mathcal{L_{\text{MWTA}}}
\end{equation}

with a hyperparameter $\gamma \in \left[0,1\right]$ to control the importance of both terms. We conduct a sensitivity study on this parameter in Appendix \textbf{D}. The \textit{meta-mode} loss is the NLL of the \textit{meta-mixture}, considering $K^{\star}\in\mathbb{N}$ as the number of \textit{meta-modes}:
\begin{align}\label{eq:meta_loss}
    \mathcal{L}_{\text{meta}} = &\frac{1}{|D|}\sum^{|D|}_{i=1} \sum^{K^{\star}}_{k^{\star}=1} - \text{log}\left(\pi_{\theta_{k^{\star}}}(X_i)\right) \nonumber \\
    &- \log \left( \mbox{Laplace}( Y_i |\ \overline{\boldsymbol{\mu}}_{\theta_{k^{\star}}}(X_i) , \overline{\mathbf{b}}_{\theta_{k^{\star}}}(X_i) ) \right)
\end{align}

This term enables backpropagating gradients to all modes, preventing hypothesis collapse and maximizing the likelihood of the meta-mixture.

$\mathcal{L_{\text{MWTA}}}$ represents the Negative Log-Likelihood (NLL) of the modes belonging to the meta-mode with the lowest error. Let $k^{*} \in [1,K^{\star}]$ denote the index of the optimal \textit{meta-mode} given $X_i$. We drop the input $X_i$ in the following notation for better legibility.
\begin{align}\label{eq:mwta_loss}
    \mathcal{L}_{\text{MWTA}} = &
    \frac{1}{|D|}\sum^{|D|}_{i=1} \frac{1}{\pi_{\theta_{k^*}}} \sum^{K'}_{k'=1} -\text{log}\big(\pi_{\theta_{k^*,k'}}\big) \nonumber \\
    &- \log \big(\mbox{Laplace}(Y_i |\ \boldsymbol{\mu}_{\theta_{k^*,k'}}, \mathbf{b}_{\theta_{k^*,k'}})\big)
\end{align}

This term encourages modes having the same \textit{meta-mode} to be close to each other as they will be optimized on the same subset of the data.
In addition, we add to both terms a classification loss such as the \textit{Cross-Entropy} loss to optimize the mixture component weights. We provide more details on the HWTA loss in Appendix \textbf{C}.

\subsection{Hierarchical Light Transformer Ensemble}

Trajectory forecasting models increasingly rely on Transformer architectures \cite{vaswani2017attention} for their ability to effectively capture spatial and temporal correlations. Ensemble methods have shown promise in enhancing the performance of these DNNs. However, like in other domains, the growing size of models poses computational challenges when ensembling. To address this, we propose leveraging the over-parameterization inherent in DNNs \cite{frankle2019lottery} to construct ensembles of lighter Transformer-based architectures while preserving their representational capacity comparable to larger models. A crucial consideration in ensembling trajectory forecasting DNNs is the linear parameter increase with the number of ensemble members. Packed Ensemble techniques \cite{laurent2023packed} have been demonstrated to be an efficient ensembling of light Convolutional Neural Networks. We focus on adapting these principles for Transformer architectures, addressing the challenge of efficiently ensembling these models. 
In particular, we propose the introduction of the Grouped Multi-head Attention Layer. Using our loss \textbf{HWTA}, we introduce two ensemble-based variants of our techniques: \textbf{HT-Ens}, a Deep Ensemble of DNNs for trajectory prediction trained with \textbf{HWTA}, and \textbf{HLT-Ens}: a lightweight version of \textbf{HT-Ens} that we developed using the Grouped Multi-head Attention Layer.

\paragraph{Grouped Fully-Connected:} 

To encapsulate several smaller fully connected layers into a larger one, we use \textit{Grouped Fully-Connected (GFC)} layers. This concept draws inspiration from the idea that grouped convolutions \cite{krizhevsky2012imagenet} enable the simultaneous training of multiple branches in parallel, as demonstrated in prior works \cite{xie2017aggregated,laurent2023packed}.
The idea is to map a portion of the input to a part of the output without overlap.

\begin{figure}[t]
\centering
    \includegraphics[width=0.5\linewidth]{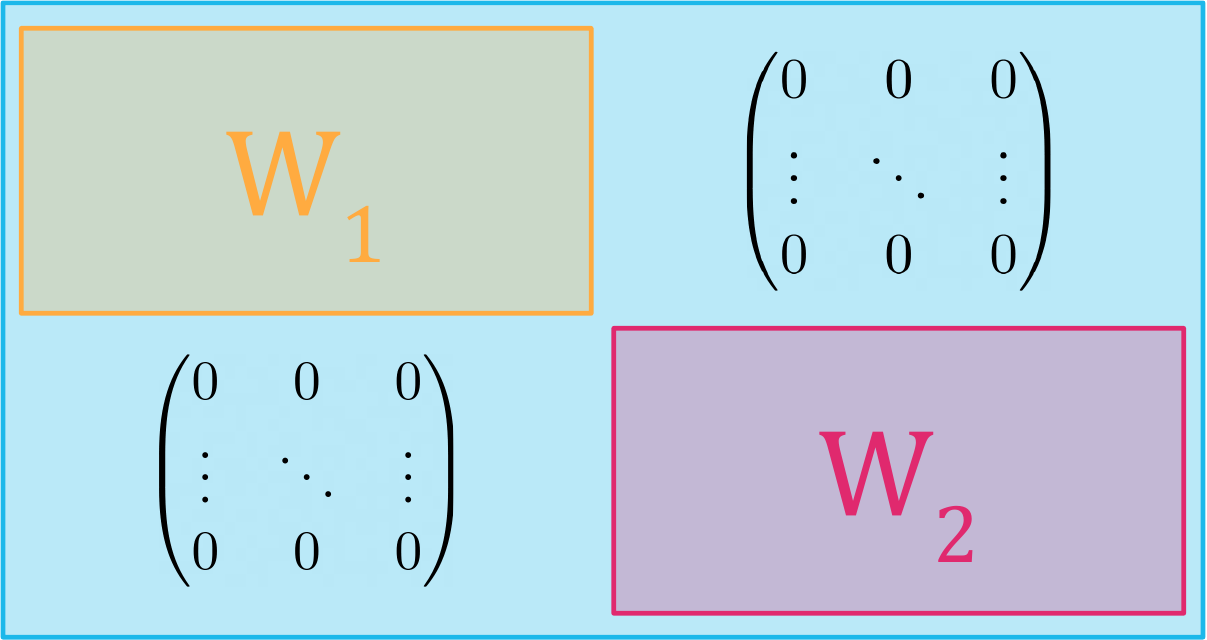} 
    \caption{\textbf{Block diagonal weight matrix for a Grouped Fully-Connected layer with $G=2$ groups.}}
    \label{fig:diag_matrix}
\end{figure}

Let us define the feature maps $\mathbf{h}^{j}\in\mathbb{R}^{D_{j}}$. Assuming we are able to split the embedding dimension $D_{j}$ into $G$ even parts, i.e., $\forall j\in\mathbb{N}$, $\mathbf{h}^{j}=\text{concat}(\mathbf{h}^{j}_{1},...,\mathbf{h}^{j}_{G})$, we create the weight tensors of a GFC layer:
\begin{equation}
   W^{j}_{g}\in\mathbb{R}^{\frac{D_j}{G}\times \frac{D_{j+1}}{G}}, \forall g\in\llbracket1,G\rrbracket
\end{equation}

\noindent
with $G\in\mathbb{N}$ the number of groups and $j\in\mathbb{N}$ the layer's index.
The output $\mathbf{h}^{j+1}$ of the grouped fully-connected operation is:
\begin{align}
    \mathbf{h}^{j+1} &= \text{GFC}(\mathbf{h}^{j};W^{j}_{1},...,W^{j}_{G}) \\
    &= \text{concat}(\mathbf{h}^{j}_{1}W^{j}_{1},...,\mathbf{h}^{j}_{G}W^{j}_{G}) \label{eq:gfc_splits}\\
    &= \mathbf{h}^{j} \text{diag}(W^{j}_{1},...,W^{j}_{G})\\
    &= \mathbf{h}^{j}W^{j}\label{eq:gfc}
\end{align}
\noindent
The transition from \cref{eq:gfc_splits} to \cref{eq:gfc} is made by the creation of a block diagonal weight matrix $W^{j}\in\mathbb{R}^{D_{j}\times D_{j+1}}$. \cref{fig:diag_matrix} illustrates this matrix for $G=2$. For memory efficiency, only $W_1$ and $W_2$ matrices are stored, dividing by $2$ the number of parameters required by the linear projection. Note that we omit the bias for brevity, as it is left unchanged.

\begin{figure}[t]
    \centering
    \begin{subfigure}[b]{0.90\linewidth}
        \centering
        \includegraphics[width=0.90\linewidth]{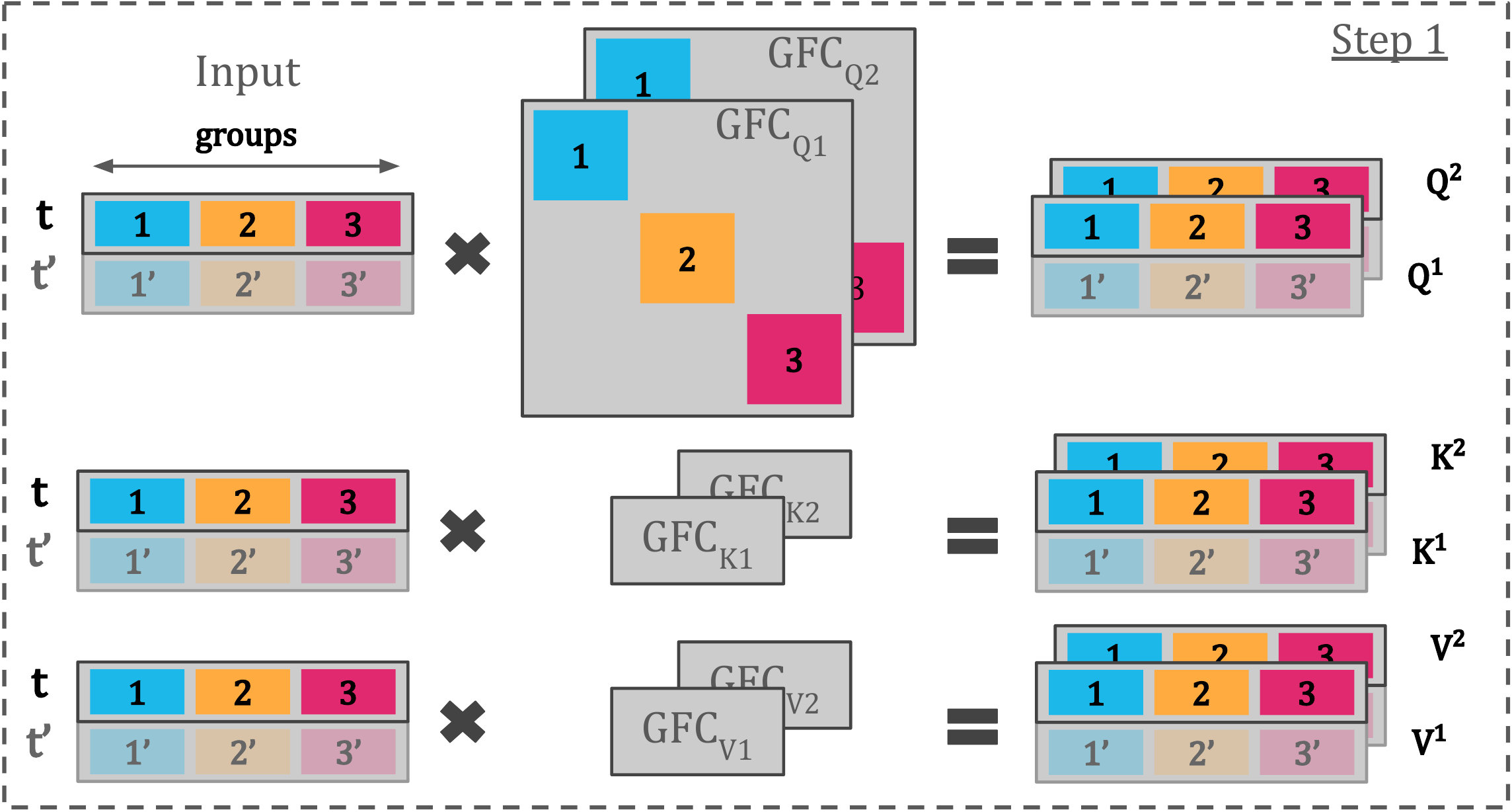} 
        \label{fig:grouped_mha_1}
    \end{subfigure}
    \begin{subfigure}[b]{0.85\linewidth}
        \centering
        \includegraphics[width=0.95\linewidth]{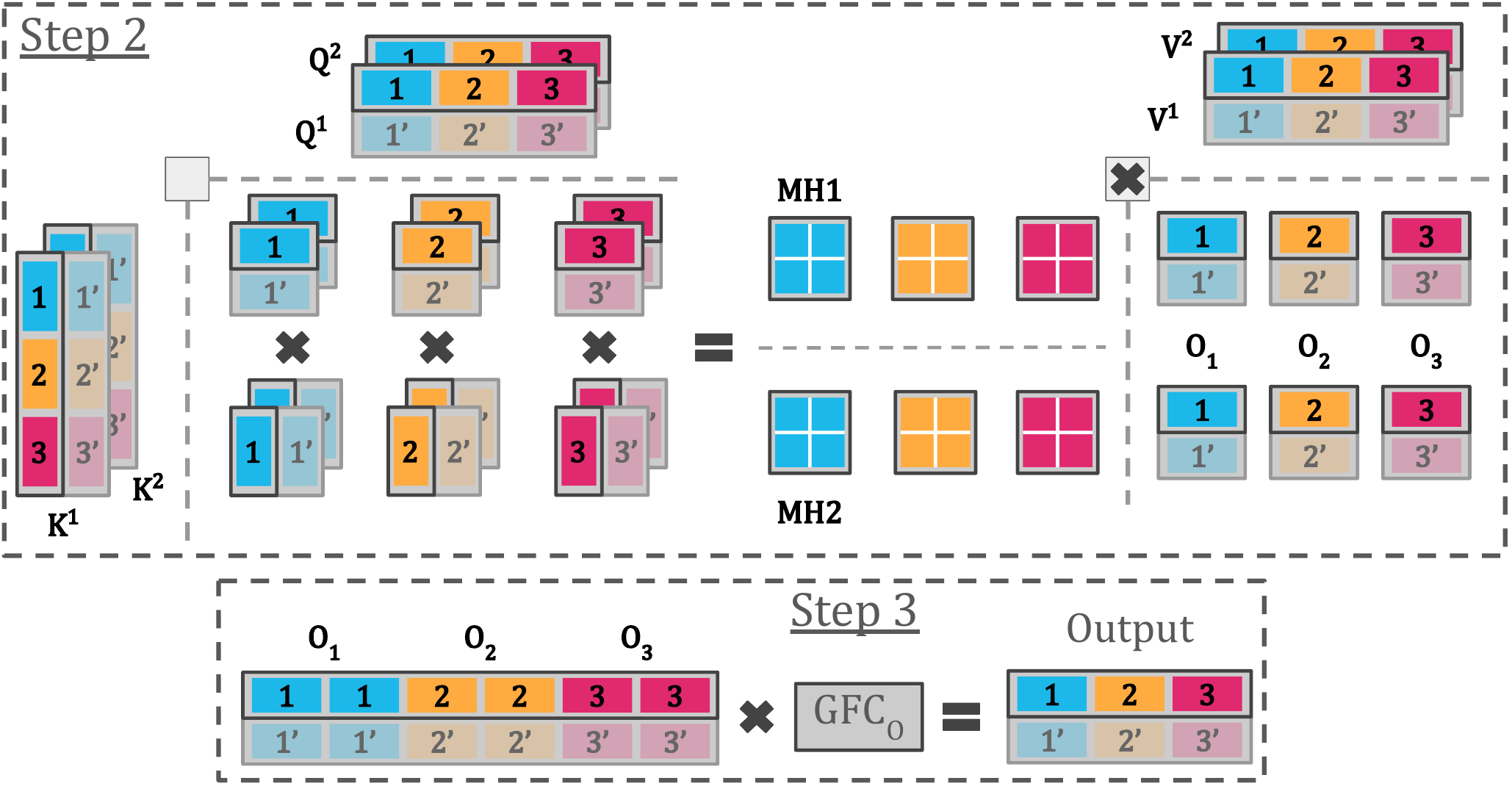} 
        \label{fig:grouped_mha_2}
    \end{subfigure}
    
    \caption{\textbf{Grouped Multi-head Attention layer operations for $H=2$ and $G=3$.} \textit{Step 1} computes the query, key, and value features for each head using grouped fully-connected layers to ensure independence between all three groups. \textit{Step 2} depicts attention mechanisms for each group in each head. \textit{Step 3} illustrates the projection of the concatenation of the heads using a grouped fully connected layer.}
    \label{fig:masks}
\end{figure}

\paragraph{Grouped Multi-head Attention:} Consider $Q\in\mathbb{R}^{n\times d_q}$ a query matrix, corresponding to $n\in\mathbb{N}$ query vectors of dimension $d_q\in\mathbb{N}$. 
Using $n_v \in \mathbb{N}$ key-value pairs with dimensions $d_v \in \mathbb{N}$, denoted as $K \in \mathbb{R}^{n_v \times d_q}$ and $V \in \mathbb{R}^{n_v \times d_v}$ respectively, the attention function is defined as follows: $\text{Att}(Q, K, V) = \text{softmax}\left(\frac{QK^{T}}{\sqrt{d_q}}\right)V$.

Vaswani \etal \cite{vaswani2017attention} extended this operation by introducing Multi-head attention. This approach simultaneously applies $H \in \mathbb{N}$ attention functions in parallel, each operating on distinct segments of the embedding dimensions. The hyperparameter $H$ can be analogously interpreted, akin to the concept of groups previously introduced, as all heads are computed in parallel. Utilizing the shared $Q$, $K$, and $V$, the multi-head attention function is expressed as:
\begin{align}
    \text{MultiHead}(Q, K, V) &= \text{concat}(O_1,...,O_H)W^{O},\\
    \text{where }O_h &= \text{Att}(QW^{Q}_{h}, KW^{K}_{h}, VW^{V}_{h}) \label{eq:head_attention}
\end{align}
\noindent
$\forall h\in\llbracket1,H\rrbracket$ and $d\in\mathbb{N}$, $W^{Q}_{h}$, $W^{K}_{h}\in\mathbb{R}^{d_q\times d^{H}_{q}}$ , $W^{V}_{h}\in\mathbb{R}^{d_v\times d_v^{H}}$, and $W^{O}\in\mathbb{R}^{d_v\times d}$ are learnable parameters. Often, in practice, $d^{H}_q=d_q/H$, $d^{H}_v=d_v/H$, and $d_q=d$. For better legibility, let us assume $d_q=d_v=d$ and use only $d$ as the embedding dimension.

We want to create a grouped version of the multi-head attention function, thus ensuring independence between subnetworks. 
Similarly to the GFC layer, we assume the embedding dimensions $d$ of $Q$, $K$, and $V$ are divisible by $G \times H$.
For the following, we will only detail the operations $Q$ as they are precisely the same for $K$ and $V$.
With $Q=\text{concat}(Q_1,...,Q_G)$, we define the $G\times H$ learnable parameters for the heads, i.e., $W^{Q}_{g,h}\in\mathbb{R}^{d_G\times d^{H}_{G}}$ where $g\in\llbracket1,G\rrbracket$, $d_G=d/G$, and $d^{H}_{G}=d/(GH)$. We apply our previously defined GFC operation on $Q$ for each head $h$ to obtain $Q'_{h}$:
\begin{align}
     Q'_{h}&= \text{GFC}(Q;W^{Q}_{1,h},...,W^{Q}_{G,h}) \\
     &= \text{concat}(Q_{1}W^{Q}_{1,h},...,Q_{G}W^{Q}_{G,h}) \label{eq:gfc_head}
\end{align}
\noindent
Similarly we define $K'_{h}$ and $V'_{h}$. Using the terms in \cref{eq:gfc_head}, we define the \textit{Grouped Multi-head Attention} (GMHA) (cf. \cref{fig:masks}) operation as:
\begin{align}
    \text{GMHA}(Q,K,V) &= \text{concat}(O^{1},...,O^{G})W^{O} \\
    \text{where } O^{g} &= \text{concat}(O^{g}_{1},...,O^{g}_{H}), \\
    \text{and } O^{g}_{h} &=  \text{Att}(Q_{g}W^{Q}_{g,h},K_{g}W^{K}_{g,h},V_{g}W^{V}_{g,h})
\end{align}
\noindent
Here, $W^{O}$ is the resulting block diagonal matrix from a GFC layer tensor weights as defined in \cref{eq:gfc}. In our architecture, the number of groups corresponds to the number of subnetworks $M$ in our ensemble. As stacking grouped operations until the end ensures the independence between gradients, we will have independent subnetworks as long as other operations, such as Layer Normalization \cite{ba2016layer}, can be grouped (\eg, Group Normalization \cite{wu2018group} instead of Layer Normalization) or have independent actions on each element in the embedding dimension.

\paragraph{Number of parameters:} By replacing fully connected and multi-head attention layers with their grouped counterpart, we reduce the size of the overall architecture. It comes down to the parameter reduction done in the GFC layer. The weight matrix encapsulates $D_{j}\times D_{j+1}$ parameters in a classic fully connected layer. On the other hand, GFC has $G$ weight matrices containing $D_{j}\times D_{j+1} \times G^{-2}$ parameters, meaning that we divide the number of parameters by the number of groups $G$. Hence, if we were to use groups to fit subnetworks within a DNN backbone, the size of the ensemble would decrease as its number of members increases. To tackle this problem, we introduce a parameter $\alpha\in\mathbb{R}$ to compensate for the effect of using groups. It can be interpreted as a width expansion factor affecting the embedding dimension size. This enables a more flexible control of the ensemble size. The sensitivity analysis corresponding to this hyper-parameter can be found in Appendix \textbf{E}.

\section{Experiments}
\label{sec:expe}

\begin{table}[t]
    \centering
    \resizebox{\linewidth}{!}
    {
    \begin{tabular}{>{\kern-\tabcolsep}lllcccccc<{\kern-\tabcolsep}}
    \toprule
        & & {\textbf{Method}} & \textbf{$\text{mADE}_1$} $\downarrow$ & \textbf{$\text{mADE}_6$} $\downarrow$  & \textbf{$\text{mFDE}_1$} $\downarrow$ & \textbf{$\text{mFDE}_6$} $\downarrow$ & \textbf{$\text{NLL}_3$} $\downarrow$ & \textbf{$\text{NLL}_6$} $\downarrow$ \\
        \midrule
        \multirow{10}{*}{\rotatebox[origin=c]{90}{\textbf{Interaction}}} & \multirow{5}{*}{\rotatebox[origin=c]{90}{\textbf{AutoBots}}} & {Legacy \cite{girgis2022latent}} & 0.67 & \second 0.27 & 1.95 & \first \textbf{0.66} & -33.70 & -60.18 \\
        & & {WTA}\cite{fan2016pointsetgenerationnetwork,rupprecht2017learning} & 0.65 & 0.28  & 1.92 & 0.70 & -33.18 & \second -68.64 \\
        & & {$\varepsilon$-WTA}\cite{rupprecht2017learning} & 0.65 & 0.28  & 1.88 & 0.70 & \second -44.72 & -68.23 \\
        & & EWTA\cite{makansi2019overcoming} & \second 0.59 & 0.28 & \second 1.72 & 0.70 & -34.85 & -65.99\\
        & & HWTA (Ours) & \first \textbf{0.57} & \first \textbf{0.26} & \first \textbf{1.69} & \second 0.68 & \first \textbf{-52.65} & \first \textbf{-69.50} \\
        \cmidrule[0.5pt](l){2-9}
        & \multirow{5}{*}{\rotatebox[origin=c]{90}{\textbf{ADAPT}}} & {Legacy}\cite{aydemir2023adapt}   & 0.71 & \first \textbf{0.30}  & 1.97 & \first \textbf{0.73} & \second -30.43 & \first \textbf{-55.14} \\
        & & {WTA}\cite{fan2016pointsetgenerationnetwork,rupprecht2017learning}  & 0.71 & \second 0.31 & 1.97 & \second 0.75 & -27.00 & \second -53.94 \\
        & & {$\varepsilon$-WTA}\cite{rupprecht2017learning}   & 0.74 & \second 0.31 & 2.05 & \second 0.75 & -28.61 & -52.93 \\
        & & EWTA\cite{makansi2019overcoming}   & \second 0.64 & 0.33 & \second 1.82 & 0.82 & \first \textbf{-33.54} & -50.99 \\
        & & HWTA (Ours) & \first \textbf{0.63} & \first \textbf{0.30} & \first \textbf{1.78} & \first \textbf{0.73} & -26.92 & -53.11 \\
        \midrule
        \multirow{10}{*}{\rotatebox[origin=c]{90}{\textbf{Argoverse 1}}} & \multirow{5}{*}{\rotatebox[origin=c]{90}{\textbf{AutoBots}}} & {Legacy\cite{girgis2022latent}} & \second 1.65 & \first \textbf{0.77} & \second 3.66 & \first \textbf{1.18} & 57.15 & 33.03 \\
        & & {WTA}\cite{fan2016pointsetgenerationnetwork,rupprecht2017learning} & 1.85 & \second 0.78  & 4.09 & \second 1.25 & 52.41 & \first \textbf{27.00} \\
        & & {$\varepsilon$-WTA}\cite{rupprecht2017learning}  & 1.76 & 0.80  & 3.87 & 1.30 & \first \textbf{50.45} & \second 27.58 \\
        & & EWTA\cite{makansi2019overcoming} & 2.22 & 0.81 & 4.87 & 1.26 & \second 51.03 & 32.26\\
        & & HWTA (Ours) & \first \textbf{1.52} & \second 0.78 & \first \textbf{3.38} & \second 1.25 & 70.05 & 32.27 \\
        \cmidrule[0.5pt](l){2-9}
        & \multirow{5}{*}{\rotatebox[origin=c]{90}{\textbf{ADAPT}}} & {Legacy}\cite{aydemir2023adapt}  & 1.89 & 0.85  & 4.27 & 1.40 & 58.02 & 34.39 \\
        & & {WTA}\cite{fan2016pointsetgenerationnetwork,rupprecht2017learning}  & 1.84 & \first \textbf{0.83} & 4.13 & \second 1.38 & 56.65 & \first \textbf{33.14} \\
        & & {$\varepsilon$-WTA}\cite{rupprecht2017learning}   & 1.89 & \first \textbf{0.83} & 4.23 & \second 1.38 & \second 55.56 & \second 33.34 \\
        & & EWTA\cite{makansi2019overcoming}   & \second 1.80 & \second 0.84 & \second 4.00 & 1.40 & 58.74 & 34.40 \\
        & & HWTA (Ours) & \first \textbf{1.61} & \second 0.84 & \first \textbf{3.58} & \first \textbf{1.37} & \first \textbf{55.10} & 35.20 \\
        \bottomrule
    \end{tabular}
    }
            \caption{
    \textbf{Loss performance comparison (averaged over $5$ runs) on Interaction and Argoverse 1 using AutoBots and ADAPT models.} We highlight the best performances in bold.}
    \label{tab:losses}
\end{table}

\begin{table}[t]
    \centering
    \resizebox{\linewidth}{!}
    {
    \begin{tabular}{>{\kern-\tabcolsep}lllccccccc<{\kern-\tabcolsep}}
    \toprule
        & & {\textbf{Strategy}} & \textbf{$\text{mADE}_1$} $\downarrow$ & \textbf{$\text{mADE}_6$} $\downarrow$  & \textbf{$\text{mFDE}_1$} $\downarrow$ & \textbf{$\text{mFDE}_6$} $\downarrow$ & \textbf{\#Prm} $\downarrow$ & \textbf{MAC} $\downarrow$ & \textbf{Time} (h) $\downarrow$ \\
        \midrule
        \multirow{12}{*}{\rotatebox[origin=c]{90}{\textbf{Interaction}}} & \multirow{6}{*}{\rotatebox[origin=c]{90}{\textbf{AutoBots}}} & top-$6$ & 0.68 & 0.37 & 1.97 & 0.99 & 3.6 & 45.66 & 1.3 \\
        & & {RIP\cite{malinin2021shifts}} & 0.68 & 0.40 & 1.96 & 1.09 & 3.6 & 45.66 & 1.3  \\
        & & {Multipath++\cite{varadarajan2022multipath++}} & 0.64 & 0.32 & 1.89 & 0.85 & 3.6 & 45.66 & 1.3  \\
        & & TENET, QCNet \cite{wang2022tenet,zhou2023query} & 0.64 & \second 0.27 & 1.89 & \second 0.67 & 3.6 & 45.66 & 1.3 \\
        & & HT-Ens. (Ours) & \first \textbf{0.57} & \first \textbf{0.25} & \first \textbf{1.69} & \first \textbf{0.62} & 3.6 & 45.66 & 1.3 \\
        & & HLT-Ens. (Ours) & \second 0.61 & \second 0.27 & \second 1.80 & 0.69 & \first \textbf{0.9} & \first \textbf{11.62} & \first \textbf{0.6}\\
        \cmidrule[0.5pt](l){2-10}
        & \multirow{6}{*}{\rotatebox[origin=c]{90}{\textbf{ADAPT}}} & top-$6$ & 0.71 & 0.36 & 1.98 & 0.92 & 4.2 & 3.81 & 2.0 \\
        & & {RIP\cite{malinin2021shifts}} & 0.75 & 0.37 & 1.84 & 0.96 & 4.2 & 3.81 & 2.0 \\
        & & {Multipath++\cite{varadarajan2022multipath++}} & 0.66 & 0.37 & 1.84 & 0.96 & 4.2 & 3.81 & 2.0\\
        & & TENET, QCNet \cite{wang2022tenet,zhou2023query} & 0.70 & \second 0.29 & 1.94 & \second 0.70 & 4.2 & 3.81 & 2.0 \\
        & & HT-Ens. (Ours) & \second 0.63 & \first \textbf{0.28} & \second 1.79 & \first \textbf{0.69} & 4.2 & 3.81 & 2.0 \\
        & & HLT-Ens. (Ours) & \first \textbf{0.56} & \second 0.29 & \first \textbf{1.62} & 0.72 & \first \textbf{2.0} & \first \textbf{0.86} & \first \textbf{0.7}\\
        \midrule
        \multirow{12}{*}{\rotatebox[origin=c]{90}{\textbf{Argoverse 1}}} & \multirow{6}{*}{\rotatebox[origin=c]{90}{\textbf{AutoBots}}} & top-$6$ & 1.63 & 0.99 & 3.61 & 1.84 & 4.5 & 60.80 & 3.3 \\
        & & {RIP\cite{malinin2021shifts}} & 1.60 & 1.05 & 3.56 & 1.98 & 4.5 & 60.80 & 3.3 \\
        & & {Multipath++\cite{varadarajan2022multipath++}} & \second 1.56 & 0.84 & \second 3.46 & 1.42 & 4.5 & 60.80 & 3.3 \\
        & & TENET, QCNet \cite{wang2022tenet,zhou2023query} & 1.62 & 0.81 & 3.59 & 1.35 & 4.5 & 60.80 & 3.3 \\
        & & HT-Ens. (Ours) & \first \textbf{1.54} & \first \textbf{0.76} & \first \textbf{3.43} & \first \textbf{1.19} & 4.5 & 60.80 & 3.3 \\
        & & HLT-Ens. (Ours) & 1.57 & \second 0.78 & 3.50 & \second 1.27 & \first \textbf{1.1} & \first \textbf{15.48} & \first \textbf{2.0} \\
        \cmidrule[0.5pt](l){2-10}
        & \multirow{6}{*}{\rotatebox[origin=c]{90}{\textbf{ADAPT}}} & top-$6$ & 1.83 & 0.95 & 4.04 & 1.72 & 4.2 & 3.81 & 5.1 \\
        & & {RIP\cite{malinin2021shifts}} & 1.64 & 1.07 & 3.68 & 2.03 & 4.2 & 3.81 & 5.1 \\
        & & {Multipath++ \cite{varadarajan2022multipath++}} & \second 1.62 & 0.98 & \second 3.61 & 1.84 & 4.2 & 3.81 & 5.1 \\
        & & TENET, QCNet \cite{wang2022tenet,zhou2023query} & 1.76 & 0.82 & 3.92 & 1.36 & 4.2 & 3.81 & 5.1 \\
        & & HT-Ens. (Ours) & 1.72 & \second 0.81 & 3.84 & \second 1.31 & 4.2 & 3.81 & 5.4 \\
        & & HLT-Ens. (Ours) & \first \textbf{1.54} & \first \textbf{0.78} & \first \textbf{3.40} & \first \textbf{1.24} & \first \textbf{2.0} & \first \textbf{0.86} & \first \textbf{2.3} \\
        \bottomrule
    \end{tabular}
    }
    \caption{
    \textbf{Ensemble strategies comparison (averaged over five runs) on Interaction and Argoverse 1 using AutoBots and ADAPT backbones.} We highlight the best performances in bold.}
        \label{tab:ensembles}
\end{table}

This section reports our experiments to validate the properties of our proposed contributions to trajectory forecasting tasks. Our evaluation focuses on assessing the accuracy of the predictions and the uncertainty estimates associated with the distribution of the resulting mixture of forecasts. We provide a performance comparison of four Transformer-based backbones to showcase the benefits of both our loss and efficient ensembling architecture on trajectory forecasting tasks. Training procedures are the same for all methods sharing the same backbone, and we report the hyperparameters in Appendix \textbf{B}. The results of Wayformer and SceneTransformer are detailed in Appendix \textbf{H}.

\paragraph{Datasets:} We evaluate our method's performance in a single agent setting on Argoverse 1 \cite{chang2019argoverse} and Interaction \cite{zhan2019interaction}. Argoverse 1 Motion Forecasting is a dataset that allows us to train a model on a vast collection of $323,557$ scenarios of $5$ seconds each. Each scenario is a 2D bird-eye-view of the scene, tracking the sampled object at $10$Hz. The Agoverse 1 challenge is to predict $3$ seconds trajectory forecasting using $2$ seconds of the past trajectory. The Interaction dataset contains road traffic data from $11$ locations recorded using drones or fixed cameras. The recordings are split into scenarios of $4$ seconds at a $10$Hz frequency. The Interaction challenge proposes to predict $3$ seconds of motion trajectory using $1$ second of prior observations.

\paragraph{Baselines:} To conduct our experiments on the Argoverse 1 and Interaction datasets, we leverage four Transformer-based backbones: AutoBots \cite{girgis2022latent}, 
ADAPT \cite{aydemir2023adapt} in \cref{tab:losses}, and SceneTransformer \cite{ngiam2021scene} and Wayformer \cite{nayakanti2023wayformer} in Appendix \textbf{H}. First, we compare our loss, HWTA, to the classical WTA \cite{rupprecht2017learning}, the relaxed WTA ($\varepsilon$-WTA)\cite{rupprecht2017learning}, and the Evolving WTA (EWTA)\cite{makansi2019overcoming} losses as well as the original losses used to train the backbones we are using (we denote these as the \textit{Legacy} loss). All methods set the number of modes to $6$ ($K=6$), and our approach considers $2$ \textit{meta-modes} ($K^{\star}=2$ and $K'=3$) unless otherwise indicated in Appendix \textbf{B}.
Then, to demonstrate the possibility of reducing the number of parameters in an ensemble of Transformer models, we report the performance of baseline ensembles (DE) with different forecast aggregation strategies: taking the top-$k$ most confident predictions, applying RIP \cite{malinin2021shifts}, the EM algorithm as in \cite{varadarajan2022multipath++}, and K-Means \cite{wang2022tenet,zhou2023query,nayakanti2023wayformer}. We compare them to the ensembling of hierarchical transformers (HT-Ens) and our hierarchical light ensembling (HLT-Ens). We post-process the outputs of methods producing more than $6$ modes using the K-Means algorithm on the final predicted positions to get $6$ trajectories.

\paragraph{Metrics:} 
We evaluate our models using standard metrics in the field \cite{chang2019argoverse,caesar2020nuscenes,wilson2021argoverse,zhan2019interaction}: minimum Average Displacement Error (\textbf{$\text{mADE}_k$}) and minimum Final Displacement Error (\textbf{$\text{mFDE}_k$}). 
Subscript $k$ denotes that these metrics are calculated over the top-$k$ most confident forecasted trajectories. We define the best trajectory as having the closest endpoint to the ground truth over the $k$ trajectories. $\text{mADE}_k$ measures the average position-wise $\ell_2$ distance between the ground-truth and the best predicted trajectory. $\text{mFDE}_k$ corresponds to the $\ell_2$ distance between the ground-truth endpoint and the best trajectory one. In addition, we propose to study the \textbf{$\text{NLL}_k$} with $k \in \{3,6\}$. We report the number of parameters (\textbf{\#Prm}) expressed in millions, the number of Mega multiply-add operations for a forward pass (\textbf{MAC}), and the training \textbf{Time} to measure ensembling approach complexity.

\begin{figure}[t]
     \centering
     \begin{subfigure}[b]{\linewidth}
         \centering
         \includegraphics[width=\textwidth]{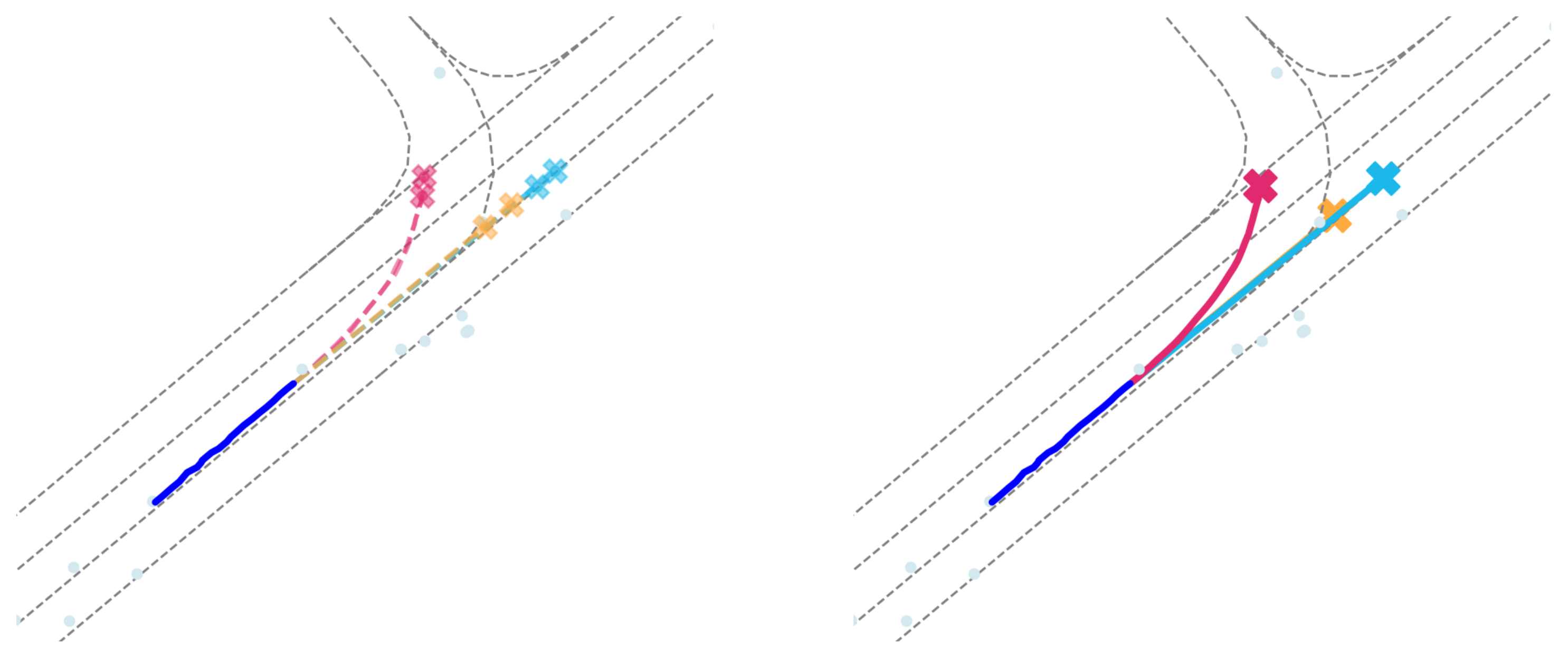}
         \caption{\textit{(left)} \textit{Sub-modes} of three \textit{meta-modes} (distinct color). \textit{(right)} \textit{meta-modes} computed as the average of their \textit{sub-modes}.}
         \label{fig:quali_0}
     \end{subfigure}
    \caption{\textbf{Examples of forecasts of an AutoBots model trained using the HWTA loss on the Argoverse 1 dataset.} It has three \textit{meta-modes} with two \textit{sub-modes} each. The observed trajectory of the agent of interest is depicted with a dark blue solid line, while the grey dashed lines are the centerlines of road lanes. We represent the model's forecasts in cyan, magenta, and yellow. Solid lines are \textit{meta-modes} and dashed lines are \textit{sub-modes}.}
    \label{fig:qualitative}
\end{figure}

\paragraph{Quantitative Results:} \cref{tab:losses} presents the average models' performance on the Argoverse 1 and the Interaction datasets over five runs. Using baseline architectures, HWTA consistently achieves attractive results compared to other optimization strategies on most metrics, especially in $\text{mADE}_1$ and $\text{mFDE}_1$. It demonstrates the ability of our loss to focus on high-probability density areas while retaining enough diversity to perform comparably with other methods.

\cref{tab:ensembles} presents the average performance of models on the Argoverse 1 and Interaction datasets over five runs using different ensemble strategies. It is shown that HT-Ens, representing a Deep Ensembles with our HWTA loss, achieves state-of-the-art performance. Furthermore, HLT-Ens, which employs our strategy to reduce the computational budget of HT-Ens, matches HT-Ens's performance while cutting the computational time by at least half. This demonstrates the significant advantages of HLT-Ens over other strategies.

On a more general note, the K-Means algorithm as an aggregation strategy produces satisfactory results for diversity. Thus, we used it to aggregate the predictions of our ensembling methods: HT-Ens and HLT-Ens.
Concerning HLT-Ens, its application to ADAPT outperforms the baseline ensemble counterpart by a large margin on $\text{mADE}_1$ and $\text{mFDE}_1$. It achieves similar results on other metrics, using a small fraction of the parameters and number of operations required by classical ensembles. The application to AutoBots also provides good results compared to DE, with much fewer parameters, but in contrast, it seems to underperform slightly compared to HT-Ens. We argue that it might be the symptom of a representation capacity too small to be subdivided into $3$ networks. In Appendix \textbf{E}, we discuss the influence of the representation capacity. In Appendix \textbf{G}, we provide more insight into the differences between HLT-Ens and DE in trajectory forecasting.

\cref{tab:stdev} demonstrates an improved stability induced by our loss in the optimization process. As WTA losses focus on the best prediction, the performance on the $\text{mADE}_6$ and $\text{mFDE}_6$ are consistent, i.e., low standard deviation across runs. However, we observe a higher performance variation on both $\text{mADE}_1$ and $\text{mFDE}_1$, suggesting instabilities in the optimization process. Concerning HWTA, in contrast, the standard deviation is much lower, demonstrating our new loss's ability to achieve consistent results across training runs.

\begin{table}[t]
    \centering
    \resizebox{\linewidth}{!}
    {
    \begin{tabular}{>{\kern-\tabcolsep}llcccccccc<{\kern-\tabcolsep}}
    \toprule
        & & \multicolumn{4}{c}{\textbf{Argoverse 1}} & \multicolumn{4}{c}{\textbf{Interaction}} \\
        & & $\text{mADE}_1$ & $\text{mADE}_6$ & $\text{mFDE}_1$ & $\text{mFDE}_6$ & $\text{mADE}_1$ & $\text{mADE}_6$ & $\text{mFDE}_1$ & $\text{mFDE}_6$ \\
        \midrule
        \multirow{4}{*}{\rotatebox[origin=c]{90}{\textbf{AutoBots}}} & WTA & 0.25 & \first \textbf{0.01} & 0.54 & \second 0.02 & \second 0.02 & \first \textbf{0.00} & \second 0.04 & \first \textbf{0.01} \\
        & $\varepsilon$-WTA & 0.22 & \second 0.02 & 0.46 & 0.04 & 0.03 & 0.01 & 0.09 & \second 0.02 \\
        & EWTA & \second 0.14 & 0.03 & \second 0.35 & 0.05 & 0.03 & 0.01 & 0.05 & \second 0.02 \\
        & HWTA & \first \textbf{0.07} & \second 0.02 & \first \textbf{0.14} & \first \textbf{0.02} & \first \textbf{0.01} & \second 0.01 & \first \textbf{0.03} & 0.03 \\
        \midrule
        \multirow{4}{*}{\rotatebox[origin=c]{90}{\textbf{ADAPT}}} & WTA & 0.14 & \first \textbf{0.03} & 0.35 & \second 0.09 & 0.04 & \second 0.02 & 0.08 & 0.05 \\
        & $\varepsilon$-WTA & \second 0.05 & \second 0.04 & \second 0.15 & 0.12 & \second 0.02 & 0.03 & \second 0.03 & \second 0.03 \\
        & EWTA & 0.12 & \second 0.04 & 0.24 & 0.10 & \second 0.02 & \second 0.02 & 0.04 & 0.04 \\
        & HWTA & \first \textbf{0.04} & \first \textbf{0.03} & \first \textbf{0.08} & \first \textbf{0.08} & \first \textbf{0.01} & \first \textbf{0.01} & \first \textbf{0.02} & \first \textbf{0.02} \\
    \bottomrule
    \end{tabular}
    }
    \caption{
    \textbf{Standard deviation on metrics over five runs.} This table reports the standard deviation of performance for WTA-based losses on the $\text{mADE}_k$ and $\text{mFDE}_k$ with $k=\{1,6\}$ metrics. Note the hyperparameters are the same as in \cref{tab:losses}.}
    \label{tab:stdev}
\end{table}

\paragraph{Qualitative Results:} \cref{fig:qualitative} illustrates our new loss behavior. As expected, the \textit{sub-modes} of the same \textit{meta-mode} are near each other. This highlights the ability of our approach to enable fast compression of the mixture distribution by taking the \textit{meta-modes} only. Doing so would retain the most diversity while simplifying the forecast. We explore this aspect in Appendix \textbf{F} and provide more qualitative results in Appendix \textbf{I}.

\section{Conclusion} \label{sec:conclusion}

In conclusion, our proposed HWTA loss and ensembling framework, HLT-Ens, introduces a novel and efficient approach to multimodal trajectory forecasting, achieving state-of-the-art under identical experimental conditions. HLT-Ens leverages a hierarchical density representation and associated loss to train Mixture Density Networks and benefits from efficient ensemble training inspired by grouped convolutions.

In light of these advancements, HLT-Ens emerges as a promising approach to structuring better multi-modal distributions, improving the performance and robustness of multi-modal trajectory forecasting systems. Yet, the computation cost of aggregation strategies for ensembles remains an obstacle to their industrial applications. We believe spatial information provided by hierarchical multi-modal predictions and parallel ensemble training could help find more efficient ways to merge forecasts.

\section{Acknowledgment}

This work was granted access to the HPC resources of IDRIS under the allocation 2023-AD011014689 made by GENCI.

{\small
\bibliographystyle{ieee_fullname}
\bibliography{egbib}
}

\newpage
\appendix

\startcontents
\renewcommand\contentsname{Table of Contents -- Supplementary Material}
{
\hypersetup{linkcolor=black}
\printcontents{ }{1}{\section*{\contentsname}}{}
}

\section{Notations}\label{ann:notations}

\cref{tab:notations} summarizes the main notations used throughout this paper.

\begin{table}[t]
    \renewcommand{\figurename}{Table}
    \renewcommand{\captionfont}{\small}
    \centering
    \resizebox{\linewidth}{!}
    {
        \begin{tabular}{@{}ll@{}}
            \toprule
            \textbf{Notations} & \textbf{Meaning} \\
            \midrule
            $\mathcal{D}=\{\mathbf{X}^{i}_{1:T}\}$ & The set of $|\mathcal{D}|$ agents' trajectories lasting $T$ time steps \\
            \midrule
            $A$ & The number of agents \\
            \midrule
            $T$ & The number of time steps \\
            \midrule
            $K$ & The number of modes, \ie, components in the mixture distribution \\
            \midrule
            $K^{\star}$ & The number of meta-modes, \ie, components in the meta-mixture distribution \\
            \midrule
            $K'$ & The number of modes within a meta-mixture component \\
            \midrule
            $a,t,k$ & The indexes of the current agent, the current time step, the current mode\\
            \midrule
            $X_i=\mathbf{X}^{i}_{1:t}$ & The observed trajectories assuming $t$ steps of context\\
            \midrule
            $Y_i=\mathbf{X}^{i}_{t+1:T}$ & The target trajectories (\ie, ground-truth of the forecasts) assuming $t$ steps of context\\
            \midrule
            $H$ & The number of attention heads in multi-head attention layers \\
            \midrule
            $M$ & The number of estimators in an ensemble, \ie, ensemble size \\
            \midrule
            $\theta_{k}$ & The set of weights of the $k$th-component of a parametric probabilistic mixture model\\
            \midrule
            $\alpha$ & The width-augmentation factor of HLT-Ens.\\
            \midrule
            $P_{\theta}$ & The probability density function of a parametric model where $\theta$ are the parameters\\
            \midrule
            $\boldsymbol{\mu}_{\theta_k}$ & The mean of the $k$th-component of a Laplace mixture distribution parametrized by $\theta$ \\
            \midrule
            $\mathbf{b}_{\theta_k}$ & The scale vector of the $k$th-component of a Laplace mixture distribution parametrized by $\theta$ \\
            \midrule
            $\boldsymbol{\pi}_{\theta}$ & The probability vector corresponding to the a mixture weights \\
            \midrule
            $\bar{\boldsymbol{\mu}}_{\theta_k}$ & The mean of the $k$th \textit{meta-mode} of a Laplace mixture distribution parametrized by $\theta$ \\
            \midrule
            $\bar{\mathbf{b}}_{\theta_k}$ & The scale vector of the $k$th \textit{meta-mode} of a Laplace mixture distribution parametrized by $\theta$ \\
            \midrule
            $\boldsymbol{\Delta}^{C}$ & The probability simplex in the $\mathbb{R}^{C}$ space\\
            \bottomrule
        \end{tabular}
    }
    \caption{\textbf{Summary of the main notations of the paper.}}
    \label{tab:notations}
\end{table}

\section{Implementation and Training Details}\label{ann:implementation}

\begin{table*}[t]
    \renewcommand{\figurename}{Table}
    \renewcommand{\captionfont}{\small}
    \centering
    \resizebox{\textwidth}{!}
    {
        \begin{tabular}{@{}llcccc@{}}
            \toprule
             & & \multicolumn{2}{c}{\textbf{Argoverse 1}} & \multicolumn{2}{c}{\textbf{Interaction}}\\
            \multirow{-2}{*}{\textbf{Parameter}}&\multirow{-2}{*}{\textbf{Description}} & \textit{AutoBots} & \textit{ADAPT} & \textit{AutoBots} & \textit{ADAPT} \\
            \midrule
            $d$ & Hidden dimension used in all model layers & 128 & 128 & 128 & 128 \\
            Batch size & Batch size during training & 128 & 128 & 128 & 128 \\
            Epochs & Number of epochs during training & 30 & 36 & 60 & 36 \\
            Learning rate & Adam Optimizer initial learning rate & 7.5e-4 & 7.5e-4 & 7.5e-4 & 7.5e-4 \\
            Decay & Multiplicative factor of learning rate decay & 0.5 & 0.15 & 0.5 & 0.15 \\ 
            Milestones & Epoch indices for learning rate decay & 5,10,15,20 & 25,32 & 10,20,30,40,50 & 25,32 \\ 
            Dropout & Dropout rate in multi-head attention layers & 0.1 & 0.1 & 0.1 & 0.1 \\
            $K$ & Number of modes of the baselines & 6 & 6 & 6 & 6 \\
            $H$ & Number of attention heads & 16 & 8 & 16 & 8 \\
            $K^{\star}$ & Number of meta-modes for our approaches & 3 & 2 & 2 & 2\\
            $K'$ & Number of modes within a meta-mode for our approaches & 2 & 3 & 3 & 3 \\
            $\alpha$ & Width factor of HLT-Ens & 1.5 & 2.0 & 1.5 & 2.0 \\
            $\gamma$ & Tradeoff of the HWTA loss & 0.6 & 0.6 & 0.6 & 0.6 \\
            \bottomrule
        \end{tabular}
    }
    \caption{\textbf{Hyperparameters summary} for both AutoBots and ADAPT backbones across all two datasets}
    \label{tab:training_details}
\end{table*}

This section details our models' implementation and training procedure for the experiments. We expect to release the corresponding code upon acceptance and validation by our industrial sponsor. We implement all networks using the PyTorch framework and train them using an Nvidia RTX A6000, except for experiments on Wayformer where a Nvidia Tesla V100 has been used.
For both datasets, we used the same simple preprocessing:
\begin{enumerate}
    \item We transform all agents and lanes coordinates into a scene-centric  view, which is centered and oriented based on the last observed state of a focal agent,
    \ie, an agent having a complete track over the scene duration.
    \item We keep only the $6$ closest agents to the origin.
    \item We keep only the $100$ closest lanes to the origin.
\end{enumerate}

\cref{tab:training_details} summarizes the hyperparameters of all our experiments. Concerning the $\varepsilon-WTA$ loss, we used $\varepsilon = 0.05$ as specified in \cite{rupprecht2017learning}. For the EWTA loss, we set up the decay over the top-$n$ modes as described in \cref{tab:ewta_milestones}.

\begin{table}[t!]
    \label{tab:ewta_milestones}
    \centering
    \resizebox{0.7\linewidth}{!}
    {
    \begin{tabular}{>{\kern-\tabcolsep}lcc<{\kern-\tabcolsep}}
    \toprule
        \textit{Backbones} & \textbf{Argoverse 1} & \textbf{Interaction}\\
        \midrule
        \textbf{AutoBots} & 5,10,15,20,25 & 10,20,30,40,50 \\
        \textbf{ADAPT} & 5,10,15,20,25 & 5,10,15,20,25 \\
    \bottomrule
    \end{tabular}
    }
    \caption{\textbf{Hyperparameter for the EWTA loss}. This table reports the epoch indices where the number of modes ($n$) to update is decremented by $1$. At initialization $n=6$.}
\end{table}

\section{HWTA Loss Details}\label{ann:loss}

\begin{figure*}
    \centering
    \begin{subfigure}[b]{0.8\textwidth}
        \centering
        \includegraphics[width=\textwidth]{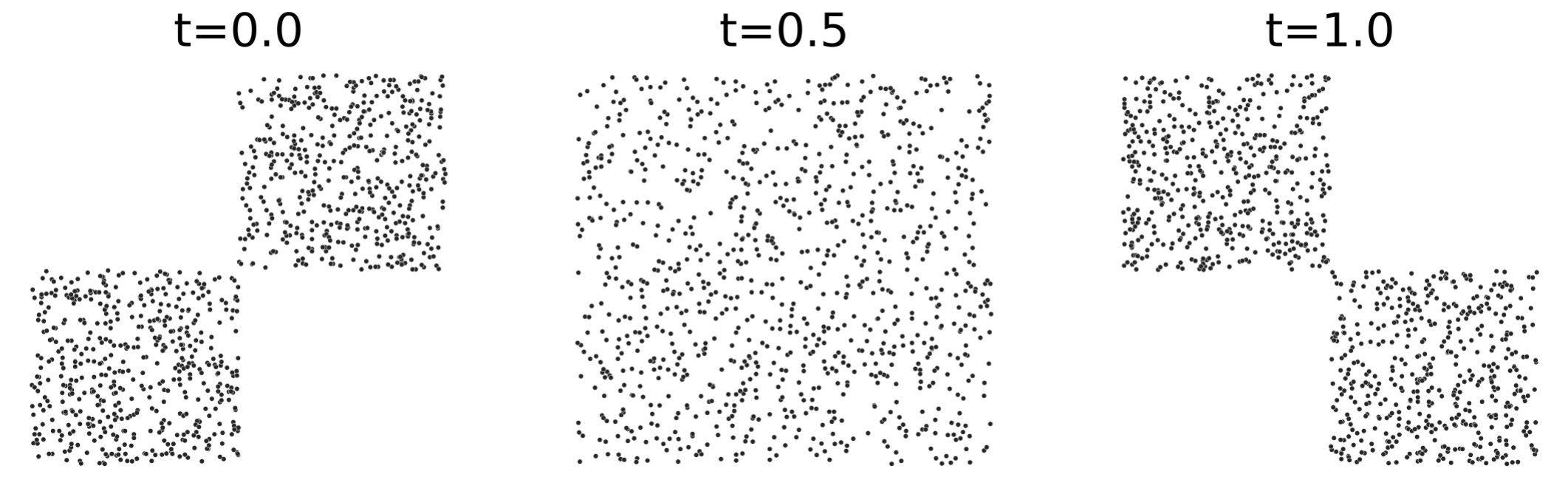} 
        \caption{Ground truth distribution}
        \vspace{2em}
        \label{fig:gt}
    \end{subfigure}
    \begin{subfigure}[b]{0.8\textwidth}
        \centering
        \includegraphics[width=\textwidth]{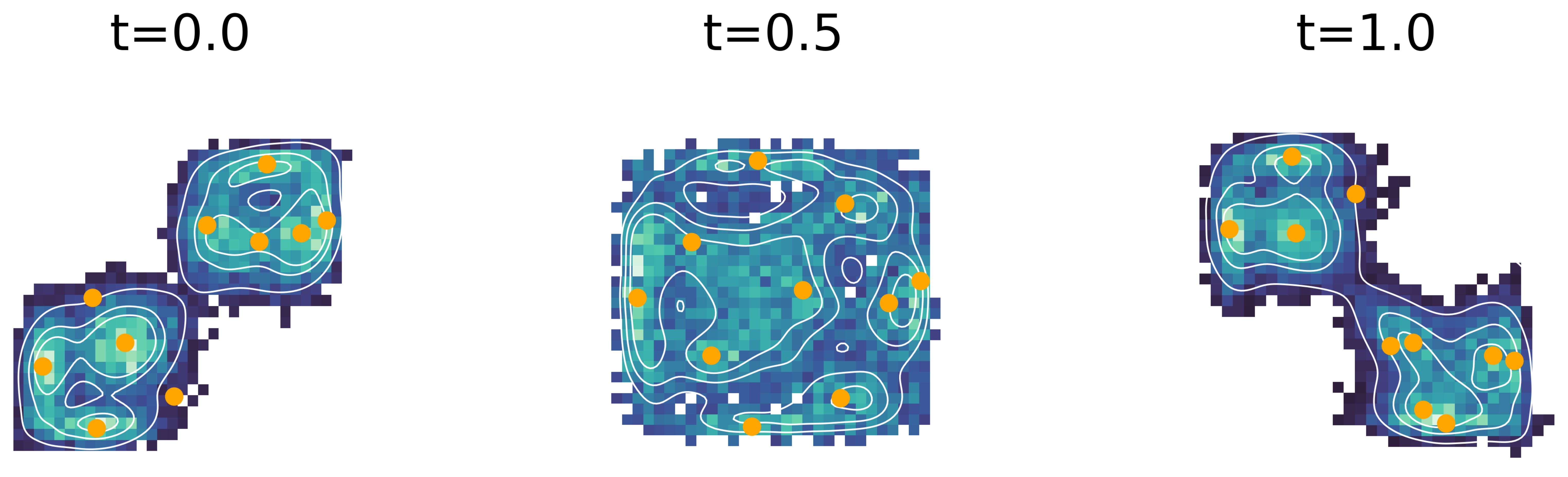} 
        \label{fig:mdn_10}
    \end{subfigure}
    \begin{subfigure}[b]{0.8\textwidth}
        \centering
        \includegraphics[width=\textwidth]{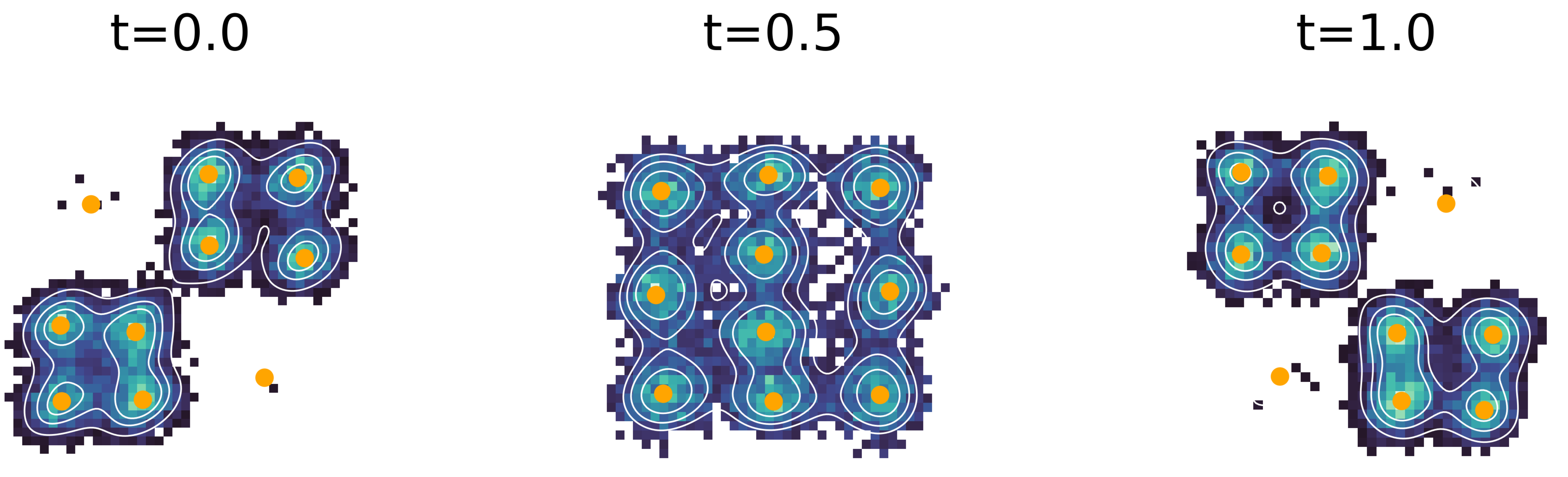} 
        \label{fig:mdn_wta_10}
    \end{subfigure}
    \begin{subfigure}[b]{0.8\textwidth}
        \centering
        \includegraphics[width=\textwidth]{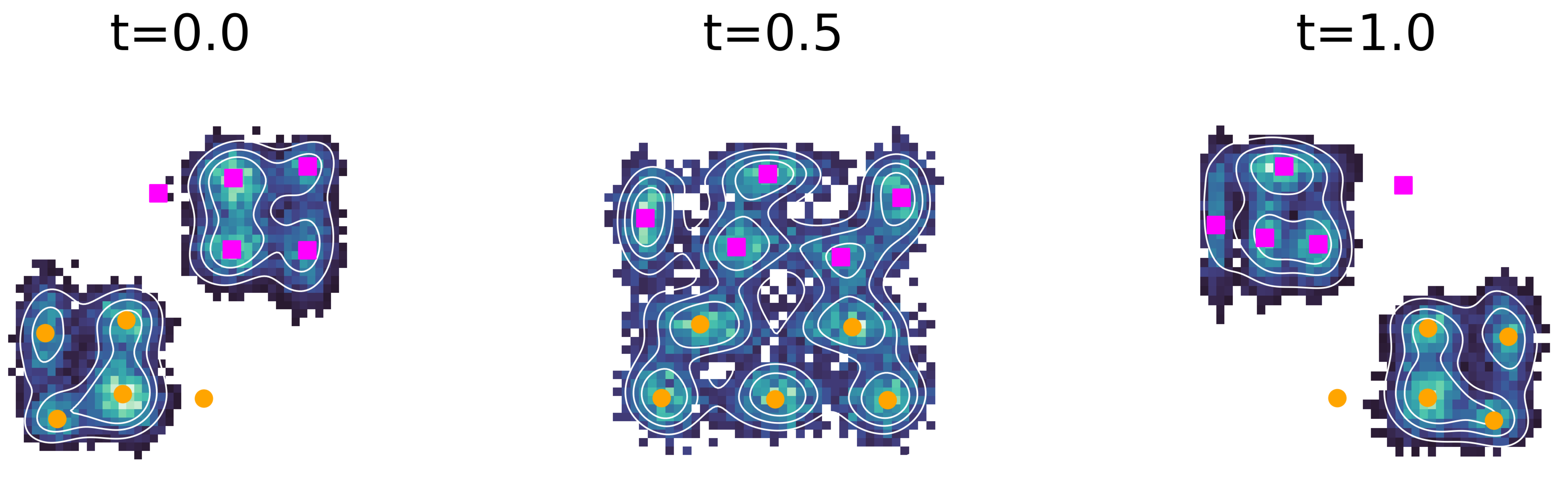} 
        \caption{Predicted distributions of small MLP models using the mixture NLL loss (\textit{top}), the WTA loss with the NLL (\textit{center}), and the HWTA loss with two meta-modes (in magenta and orange) (\textit{bottom})}
        \label{fig:hwta_10}
    \end{subfigure}
    \caption{\textbf{Temporal 2D distributions}}
    \label{fig:2d_toy}
\end{figure*}

This section provides more explanations and insights on the HWTA loss. First, we present how classification loss terms are encompassed within the \textit{meta-mode} and MWTA losses. Then, we illustrate our loss behavior on a 2D toy dataset (\cref{fig:2d_toy}).

Sec. \textbf{\textcolor{red}{4}} defines both terms in HWTA loss as the NLL loss of the meta-mixture and the best meta-mode mixture, respectively. Yet, in practice, these terms are combined with a classification loss. Indeed, we leverage the loss formulation in \cite{girgis2022latent} to define $\mathcal{L}^{\mathcal{C}}_{meta}$ and $\mathcal{L}^{\mathcal{C}}_{\text{MWTA}}$:
\begin{align}
    \mathcal{L}^{\mathcal{C}}_{meta} &= \frac{1}{K} \sum_{Z} D_{\text{KL}}(Q(Z)||P_{\theta}(Z | Y, X))\\
    \mathcal{L}^{\mathcal{C}}_{\text{MWTA}} &= \frac{1}{K'} \sum_{Z'} D_{\text{KL}}(Q(Z')||P_{\theta}(Z' | Y, X)),
\end{align}

where $Z$ and $Z'$ are discrete latent variables corresponding to the meta-modes and their respective modes. $D_{\text{KL}}(.||.)$ is the Kullback-Leibler divergence between the approximated posterior and the actual posterior. As in \cite{girgis2022latent}, we set $Q(Z)=P_{\theta_{\text{old}}}(Z|Y,X)$ and $Q(Z')=P_{\theta_{\text{old}}}(Z'|Y,X)$, where $\theta_{\text{old}}$ are the parameters before the optimization step. %
Eq. (\textcolor{red}{7}) 
becomes:
\begin{equation}\label{eq:hwta_full}
    \mathcal{L} = \gamma \times (\mathcal{L_{\text{meta}}} + \mathcal{L}^{\mathcal{C}}_{meta}) + (1-\gamma)\times (\mathcal{L_{\text{MWTA}}}+\mathcal{L}^{\mathcal{C}}_{\text{MWTA}})
\end{equation}

Inspired by \cite{rupprecht2017learning}, we utilize their custom toy dataset, which comprises a two-dimensional distribution evolving over time $t\in\left[0,1\right]$. They achieve this by dividing a zero-centered square into $4$ equal regions and transitioning from having high probability mass in the lower-left and top-right quadrants to having high probability mass in the upper-left and lower-right ones. 
Following their notation, the sections are defined as:

\begin{align}
    S_1 &= \left[-1, 0\right[ \times \left[-1, 0\right[ \subset \mathbb{R}^2\\
    S_2 &= \left[-1, 0\right[ \times \left[0, 1\right] \subset \mathbb{R}^2\\
    S_3 &= \left[0, 1\right] \times \left[-1, 0\right[ \subset \mathbb{R}^2\\
    S_4 &= \left[0, 1\right] \times \left[0, 1\right] \subset \mathbb{R}^2\\
    S_5 &= \mathbb{R}^2 \backslash \{S_1\cup S_2\cup S_3\cup S_4\}
\end{align}

and their respective probabilities being $P(S_1)=P(S_4)=\frac{1-t}{2}$, $P(S_2)=P(S_3)=\frac{t}{2}$, and $P(S_5)=0$. Whenever a region is selected, a point is sampled from it uniformly. \cref{fig:gt} illustrates such distribution for $t\in\{0, 0.5, 1\}$. To better illustrate the loss dynamics, we train a basic three-layer fully connected network with $50$ neurons in each hidden layer and ReLU activation function, similar to \cite{rupprecht2017learning}.
Given the time $t$, we are interested in modeling a two-dimensional distribution using $K=10$ modes. We visually compare the effect of the mixture NLL loss, the WTA loss combined with the NLL loss, and our HWTA loss (assuming $2$ meta-modes here) in \cref{fig:hwta_10}. Although the NLL loss seems to capture the underlying distribution better, the mode diversity is compromised by some redundant modes, diminishing their coverage. The WTA loss on the mixture NLL creates a Voronoï tesselation of the space, \ie, the modes are efficiently placed for coverage. Yet, it provides no information on what mode to keep if we subsequently reduce the number of modes.
On the contrary, the hierarchy in our loss enables us to give $2$ levels of modeling. One can directly take the meta-modes to maximize coverage and diminish the overall number of modes. We illustrate the benefit of this strategy in \cref{ann:robustness}.

\section{Loss Parameter Sensitivity Study}\label{ann:loss_sensitivity}

\begin{figure*}[t]
\centering
    \includegraphics[width=\textwidth]{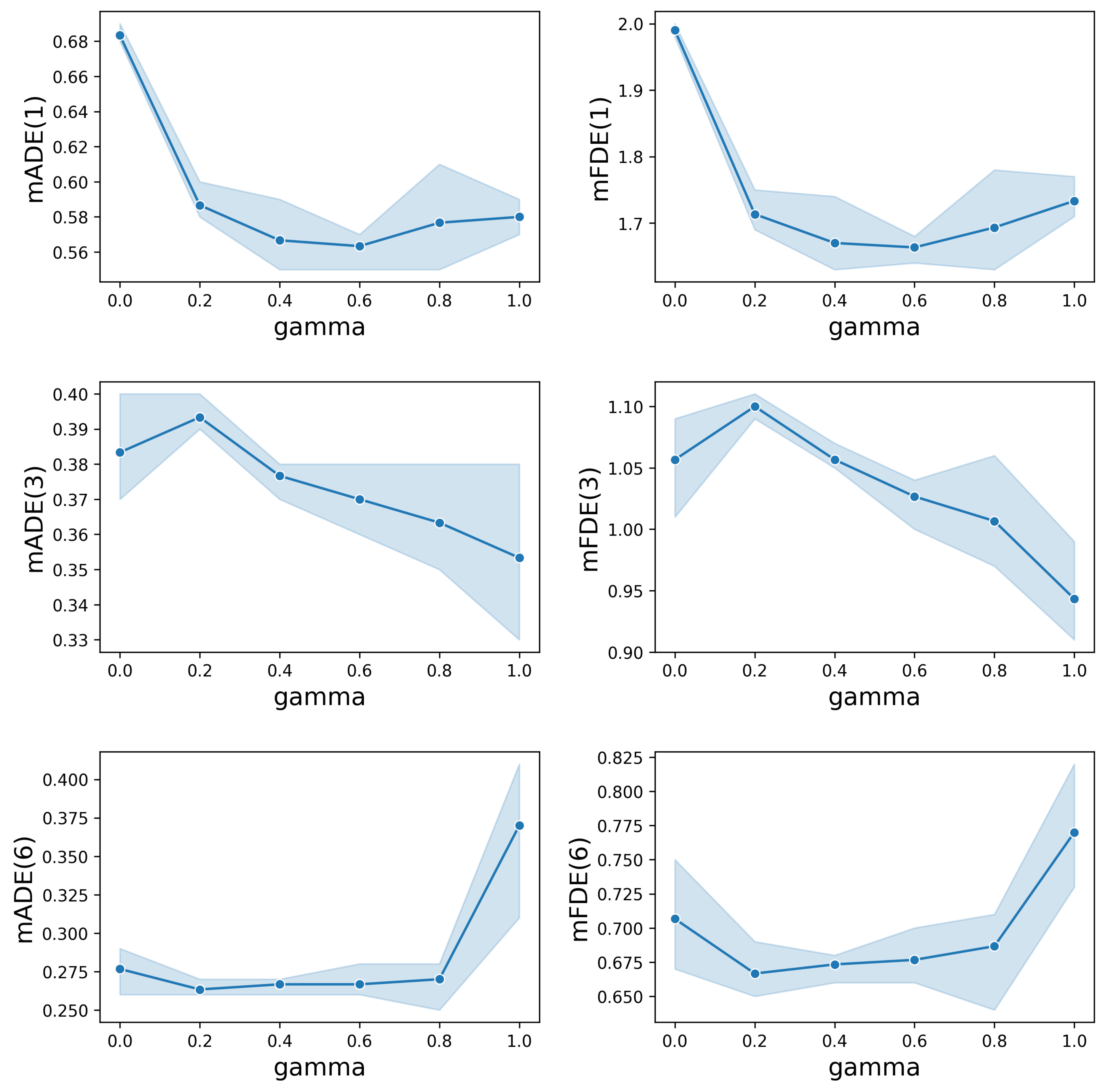} 
    \caption{\textbf{Sensitivity analysis on $\gamma$ (gamma)} using AutoBots backbone on the Interaction dataset.}
    \label{fig:tradeoff}
\end{figure*}

This section showcases the effect of $\gamma$ on our HWTA loss $\mathcal{L}$. In particular, \cref{fig:tradeoff} illustrates the performance variation for different $\gamma$ values. Note that for $\gamma=0.0$ the \textit{meta-mode} loss $\mathcal{L}_{\text{meta}}$ is not used and for $\gamma=1.0$ the loss $\mathcal{L}_{\text{MWTA}}$ is annealed. From our experiments, $\mathcal{L}_{\text{meta}}$ is necessary for better accuracy on most confident predictions as we observe a dramatic decrease in $\text{mADE}_{1}$ and $\text{mFDE}_{1}$

\section{Importance of the Width Factor}\label{ann:width_factor}

\begin{table}[t]
    \centering
    \resizebox{0.9\linewidth}{!}
    {
    \begin{tabular}{@{}c|cccc|c<{\kern-\tabcolsep}}
    \toprule
        $\boldsymbol{\alpha}$ value & \textbf{$\text{mADE}_{1}$} $\downarrow$ & \textbf{$\text{mFDE}_{1}$} $\downarrow$ & \textbf{$\text{mADE}_{6}$} $\downarrow$ & \textbf{$\text{mFDE}_{6}$} $\downarrow$ & \textbf{\#Prm} (M) \\
        \midrule
        1 & 0.78 & 2.24 & 0.45 & 1.18 & 0.4 \\
        2 & 0.61 & 1.79 & 0.34 & 0.87 & 1.5 \\
        3 & 0.55 & 1.58 & 0.28 & 0.69 & 3.1 \\
        4 & 0.51 & 1.48 & 0.25 & 0.62 & 5.5 \\ 
    \bottomrule
    \end{tabular}
    }
    \caption{\textbf{Performance of HLT-Ens - ADAPT (averaged over three runs) on Interaction wrt.} $\boldsymbol{\alpha}$. Our ensemble has $M=4$ subnetworks, with $K=2$ meta-modes containing $K'=3$ modes each.}
    \label{tab:alpha_sensi1}
\end{table}

\begin{table}[t]
    \centering
    \resizebox{0.9\linewidth}{!}
    {
    \begin{tabular}{@{}c|cccc|c<{\kern-\tabcolsep}}
    \toprule
        $\boldsymbol{\alpha}$ value & \textbf{$\text{mADE}_{1}$} $\downarrow$ & \textbf{$\text{mFDE}_{1}$} $\downarrow$ & \textbf{$\text{mADE}_{6}$} $\downarrow$ & \textbf{$\text{mFDE}_{6}$} $\downarrow$ & \textbf{\#Prm} (M) \\
        \midrule
        3 & 0.62 & 1.82 & 0.36 & 0.93 & 1.7 \\
        4 & 0.59 & 1.71 & 0.32 & 0.82 & 2.9 \\
        6 & 0.53 & 1.53 & 0.27 & 0.69 & 6.3 \\
        8 & 0.51 & 1.47 & 0.26 & 0.65 & 10.9 \\
    \bottomrule
    \end{tabular}
    }
    \caption{\textbf{Performance of HLT-Ens - ADAPT (averaged over three runs) on Interaction wrt.} $\boldsymbol{\alpha}$. Our ensemble has $M=8$ subnetworks, with $K=2$ meta-modes containing $K'=3$ modes each.}
    \label{tab:alpha_sensi2}
\end{table}

Our efficient ensembling architecture depends on two hyperparameters. $M$ corresponds to the ensemble size, and $\alpha$, as a factor on the embedding size, controls the width of the DNN. We evaluate the sensitivity of HLT-Ens to these parameters by training models using the ADAPT backbone on the Interaction dataset with various settings. \cref{tab:alpha_sensi1} and \cref{tab:alpha_sensi2} showcase the effect of $\alpha$ for $4$ and $8$ subnetworks respectively. Increasing $\alpha$ enables better results until it reaches a plateau at the cost of more parameters.

\section{Robustness Analysis over Mode Number}\label{ann:robustness}

\begin{figure*}[t!]
\centering
    \includegraphics[width=0.8\textwidth]{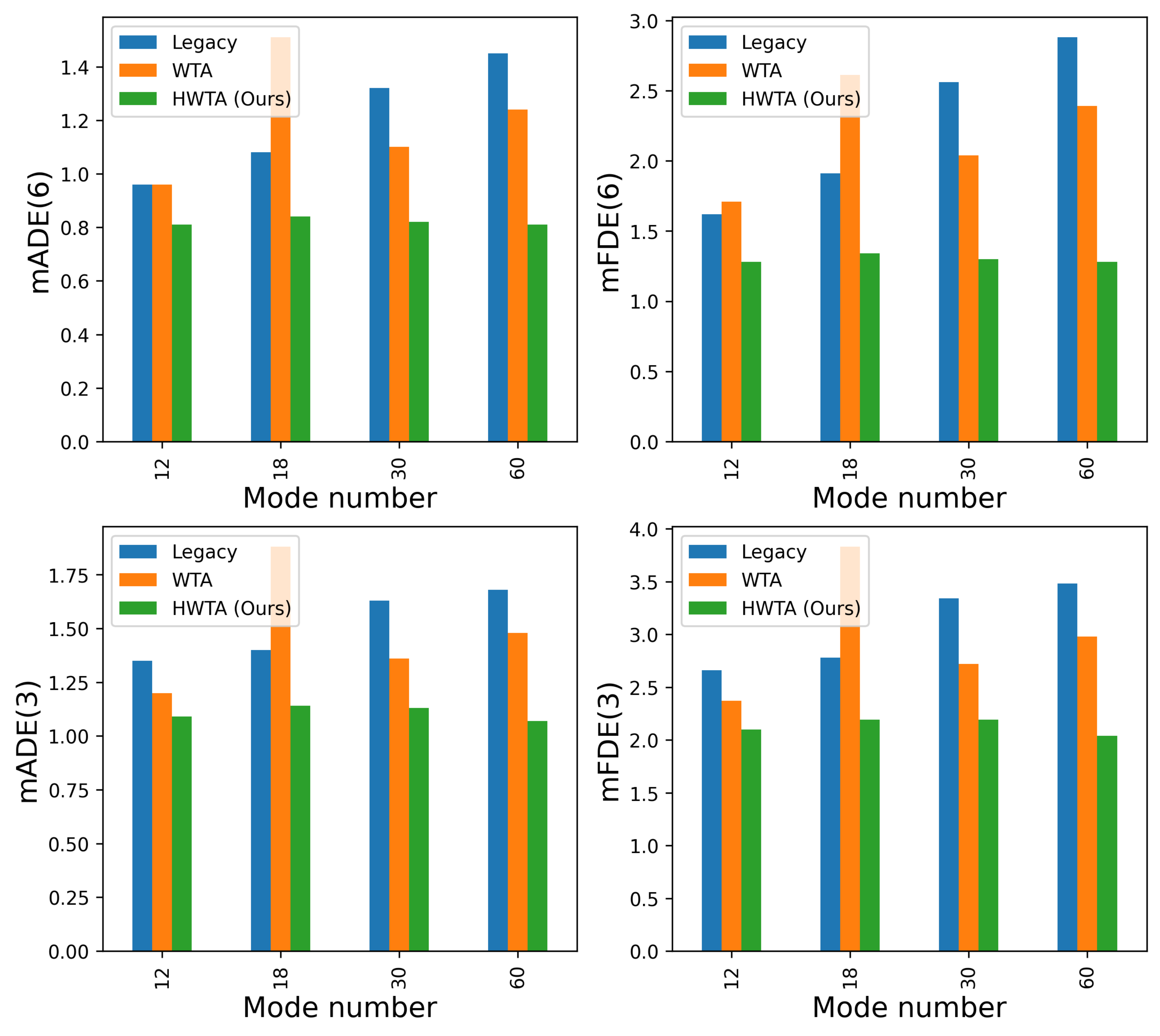}
    \caption{\textbf{Performance comparison under variations of the number of modes.} We train three AutoBots using the \textit{Legacy} loss, the WTA loss, and the HWTA loss on Argoverse 1. We report their performance in terms of $\text{mADE}_3$, $\text{mFDE}_3$, $\text{mADE}_6$ and $\text{mFDE}_6$. For our loss, we considered $6$ \textit{meta-modes}.}
    \label{fig:robustness}
\end{figure*}

The number of modes is a critical hyperparameter significantly impacting the model's performance. Predicting more modes and choosing the most confident ones is often preferred to cover the true multimodal distribution. Yet, doing so might decrease the diversity in the predicted modes as the most confident mode is likely to have duplicates.
Using hierarchy in the mixture distribution, we showcase that we can reduce our mixture complexity while retaining most of its diversity. \cref{fig:robustness} illustrates the impact of the number of modes over the performance on diversity metrics (\ie, $\text{mADE}_6$ and $\text{mFDE}_6$). The comparison is done with three different settings (\textit{Legacy} loss, WTA loss, and HWTA loss), all using AutoBots backbone and trained on Argoverse 1. Concerning our loss, instead of taking the $6$ most confident trajectories, we use the $6$ \textit{meta-modes} as a set of forecasts. Doing so seems to stabilize the performance, suggesting our loss enables more robustness to variations in the number of modes with a fixed number of \textit{meta-modes}.

\section{Diversity in Ensembles of Mixtures}

\begin{figure}
    \centering
    \begin{subfigure}[b]{0.45\linewidth}
        \centering
        \includegraphics[width=\linewidth]{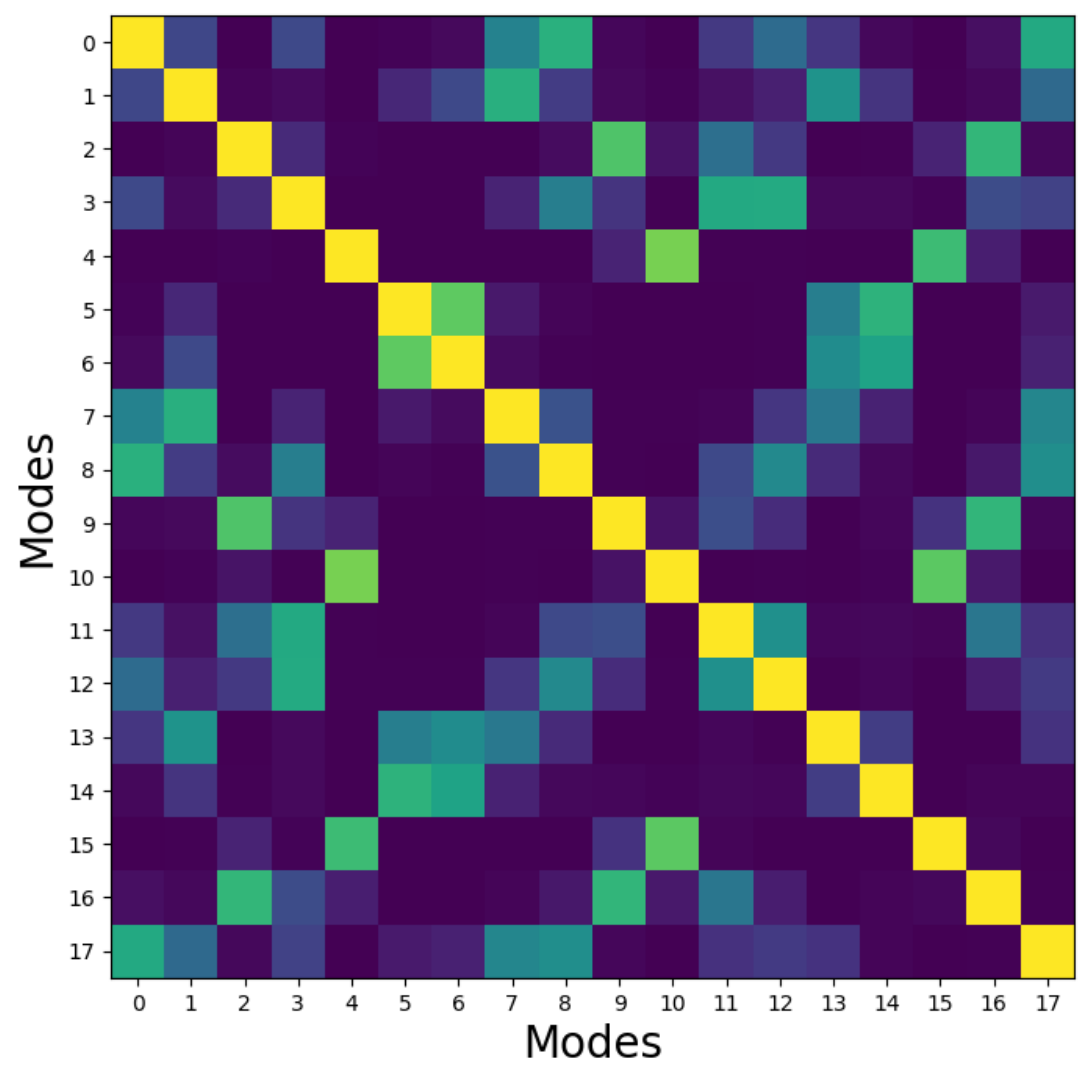} 
        \caption{DE similarity matrix}
        \label{fig:cluster}
    \end{subfigure}
    \begin{subfigure}[b]{0.45\linewidth}
        \centering
        \includegraphics[width=\linewidth]{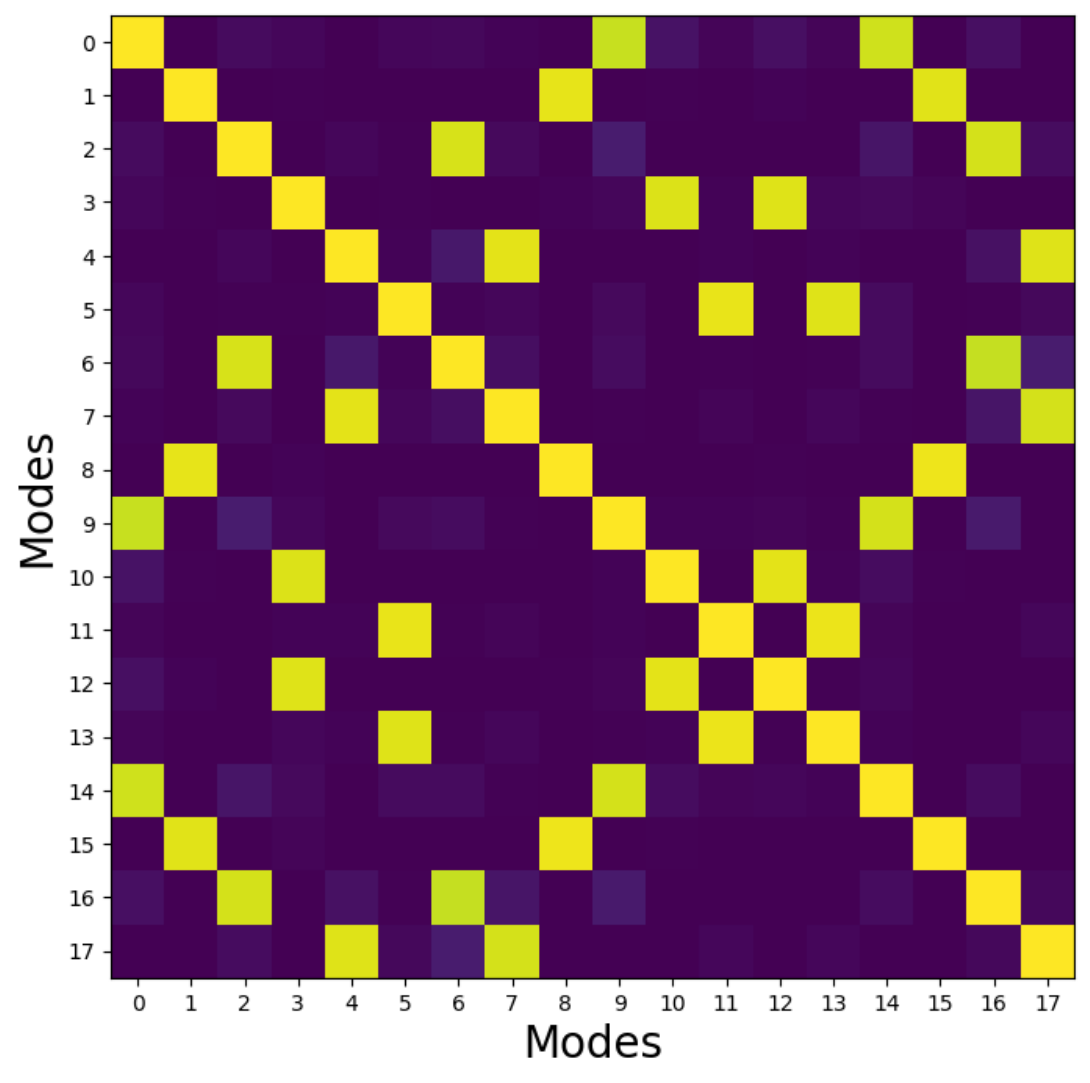} 
        \caption{HLT-Ens similarity matrix}
        \label{fig:pe_cluster}
    \end{subfigure}
    \caption{\textbf{Similarity matrices for DE and HLT-Ens on Argoverse 1.} We use the AutoBots backbone to construct ensembles of size $M=3$. For HLT-Ens, we set $\alpha=1.5$.}
    \label{fig:diversity}
\end{figure}

Prediction diversity is essential for the performance of ensembles. In \cite{laurent2023packed}, the authors present two sources of stochasticity in the training process producing diversity among ensemble members: the random initialization of the model's parameters and the shuffling of the batches. HLT-Ens does not benefit from this last source of stochasticity, yet it has comparable performance. To provide more insight into the effect on the diversity of this source of stochasticity in the training of trajectory forecasting models, we conducted a small experiment on the consistency of the cluster found by the \textit{post-hoc} KMeans algorithm on our ensembles. \cref{fig:diversity} presents the similarity matrices for both DE and HLT-Ens. Each cell represents the rate of two modes being clusterized in the same cluster by the KMeans algorithm on the validation set of Argoverse 1. The clustering appears more consistent for HLT-Ens (i.e., the similarity matrix is sparser), highlighting the possibility of clustering only once the modes and applying it without too much performance loss compared to executing a KMeans algorithm for each sample. 

\section{Wayformer and SceneTransformer Experiments}

\begin{table}[t]
    \label{tab:scene_transformer}
    \centering
    \resizebox{\linewidth}{!}
    {
    \begin{tabular}{>{\kern-\tabcolsep}llcccccccc<{\kern-\tabcolsep}}
    \toprule
        &{\textbf{Method}} & \textbf{$\text{mADE}_1$} $\downarrow$ & \textbf{$\text{mADE}_6$} $\downarrow$  & \textbf{$\text{mFDE}_1$} $\downarrow$ & \textbf{$\text{mFDE}_6$} $\downarrow$ & \textbf{$\text{NLL}_3$} $\downarrow$ & \textbf{$\text{NLL}_6$} $\downarrow$ & \textbf{\#Prm} $\downarrow$\\
        \midrule
        \multirow{7}{*}{\rotatebox[origin=c]{90}{\textbf{SceneTransformer}}}&\multicolumn{7}{c}{\tablesep \textbf{Single model}}\\
        &{Legacy} & 0.66 & \first \textbf{0.21} & 1.50 & \first \textbf{0.39} & 18.57 & 17.23 & \second \\
        &HWTA (Ours) & \first \textbf{0.48} & 0.26 & \first \textbf{1.17} & 0.52 & \first \textbf{-4.74} & \first \textbf{-11.15} & \multirow{-2}{*}{\second 11.8}\\
        &\multicolumn{7}{c}{\tablesep \textbf{Ensemble}}\\
        &DE  & 0.57 & \first \textbf{0.21} & 1.30 & \first \textbf{0.60} & 23.09 & 22.59 & 35.4\\
        &HT-Ens (Ours) & \second 0.50 & \second 0.23 & \second 1.19 & \second 0.47 & \first \textbf{-7.34} & \first \textbf{-9.94} & 35.4\\
        &HLT-Ens (Ours) & \first \textbf{0.49} & 0.25 & \first \textbf{1.17} & 0.53 & \second -4.24 & \second -9.59 & \first \textbf{8.9}\\
        \midrule
        \multirow{7}{*}{\rotatebox[origin=c]{90}{\textbf{Wayformer}}}&\multicolumn{7}{c}{\tablesep \textbf{Single model}}\\
        &{WTA} & 1.75 & 0.51 & 4.13 & 1.20 & 49.92 & -11.58 & \first \\
        &HWTA (Ours) & \first \textbf{0.81} & \first \textbf{0.40} & \first \textbf{2.24} & \first \textbf{0.95} & \first \textbf{-15.54} & \first \textbf{-49.71} & \multirow{-2}{*}{\first \textbf{1.1}}\\
        &\multicolumn{7}{c}{\tablesep \textbf{Ensemble}}\\
        &DE  & 1.01 & 0.48 & 2.74 & 1.19 & -12.81 & -21.23 & 3.3\\
        &HT-Ens (Ours) & \second 0.74 & \second 0.35 & \second 2.06 & \second 0.80 & \second -25.80 & \second -47.36 & 3.3\\
        &HLT-Ens (Ours) & \first \textbf{0.68} & \first \textbf{0.33} & \first \textbf{1.87} & \first \textbf{0.71} & \first \textbf{-27.64} & \first \textbf{-49.78} & \second 1.5\\
    \bottomrule
    \end{tabular}
    }
    \caption{
    \textbf{Performance comparison on Interaction for the SceneTransformer and Wayformer backbones.} All ensembles have $M=3$ subnetworks and are followed by a KMeans algorithm to form $6$ trajectory clusters from which we take the centroids; we highlight the best performances in bold. For our method, we consider $\alpha=1.5$ for SceneTransformer and $\alpha=2.0$ for Wayformer. The number of parameters is expressed in millions.}
\end{table}

\cref{tab:scene_transformer} presents the performance of our method on the SceneTransformer and Wayformer backbones trained on the Interaction dataset. These results are based on custom re-implementations we developed for both methods. These preliminary results showcase the ability of our approach to improve the most confident forecast accuracy and the quality of the predicted multimodal distribution compared to the original loss presented in both papers. Our method applied to Wayformer outperforms its counterpart on all metrics. Interestingly, \textbf{HLT-Ens} even outperforms \textbf{HT-Ens}, making it an attractive choice. It only has slightly more parameters than the classic Wayformer.
Concerning SceneTransformer, we observe a lack of diversity in our predicted modes. We argue it might be necessary to try out other values of $\gamma$ (we used $\gamma=0.6$ here).

\section{Additional Qualitative Results}

\begin{figure*}
    \centering
    \begin{subfigure}[b]{0.7\linewidth}
        \centering
        \includegraphics[width=\linewidth]{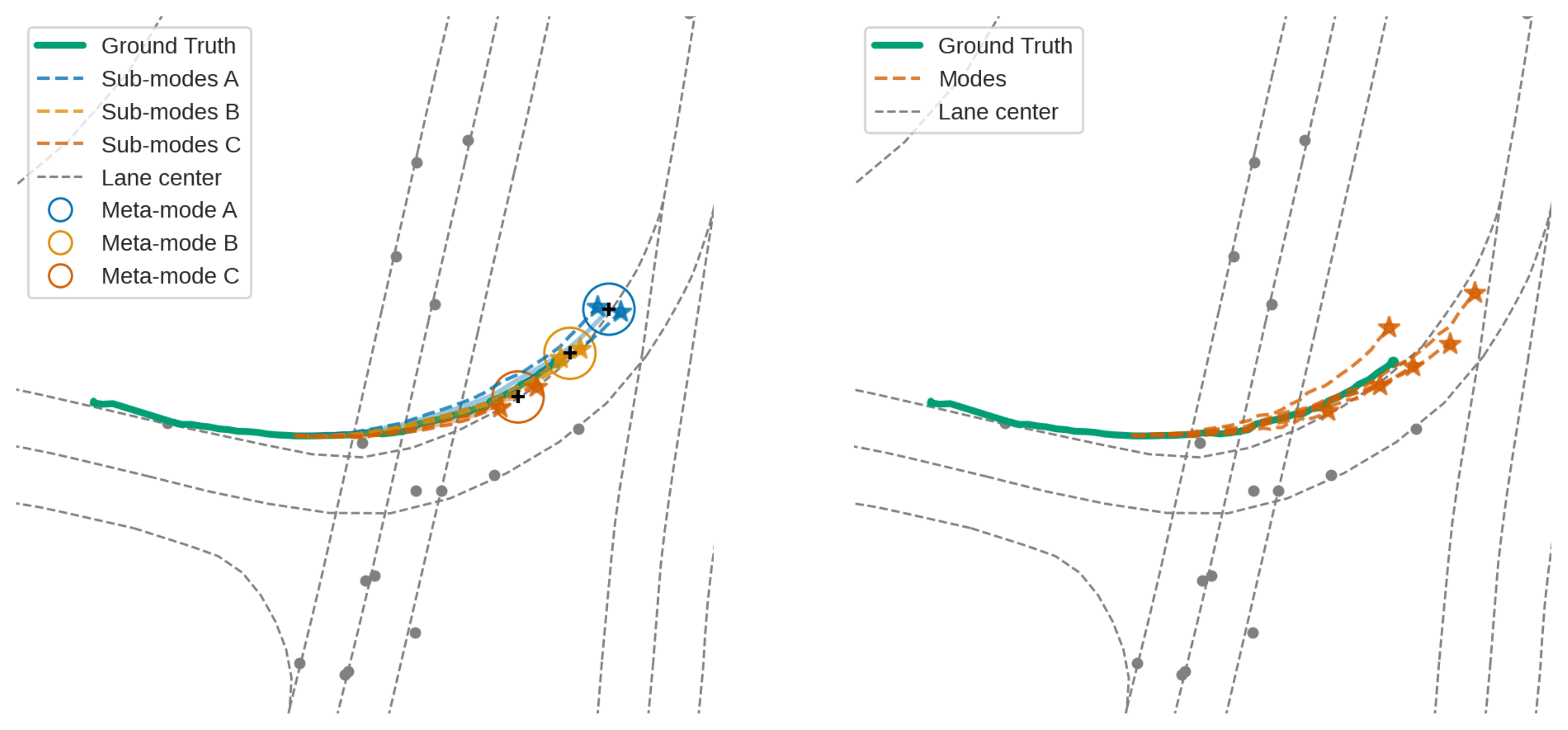} 
        \caption{Left turn scenario}
        \label{fig:quali_1}
    \end{subfigure}
    \begin{subfigure}[b]{0.7\linewidth}
        \centering
        \includegraphics[width=\linewidth]{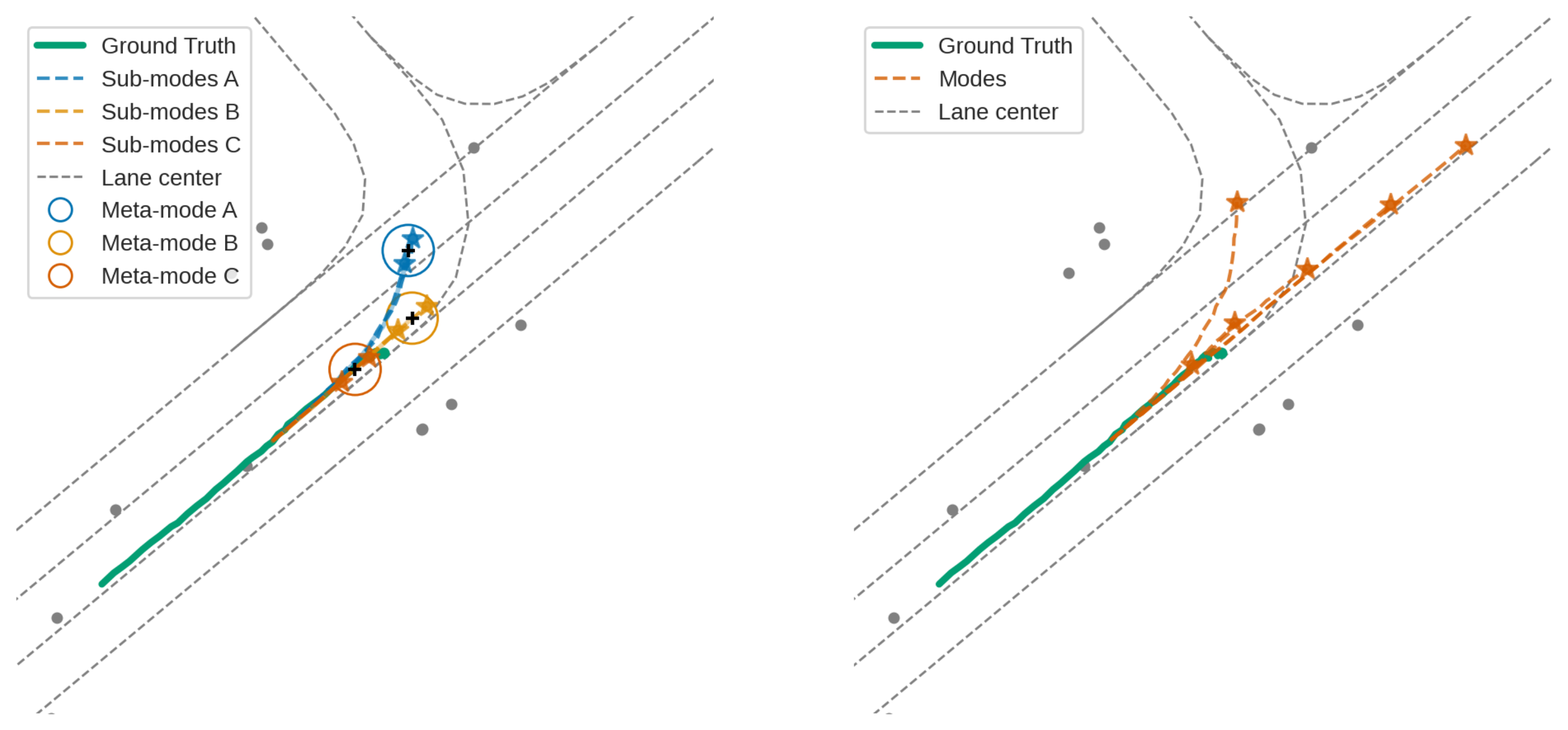} 
        \caption{Intersection scenario}
        \label{fig:quali_2}
    \end{subfigure}
    \begin{subfigure}[b]{0.7\linewidth}
        \centering
        \includegraphics[width=\linewidth]{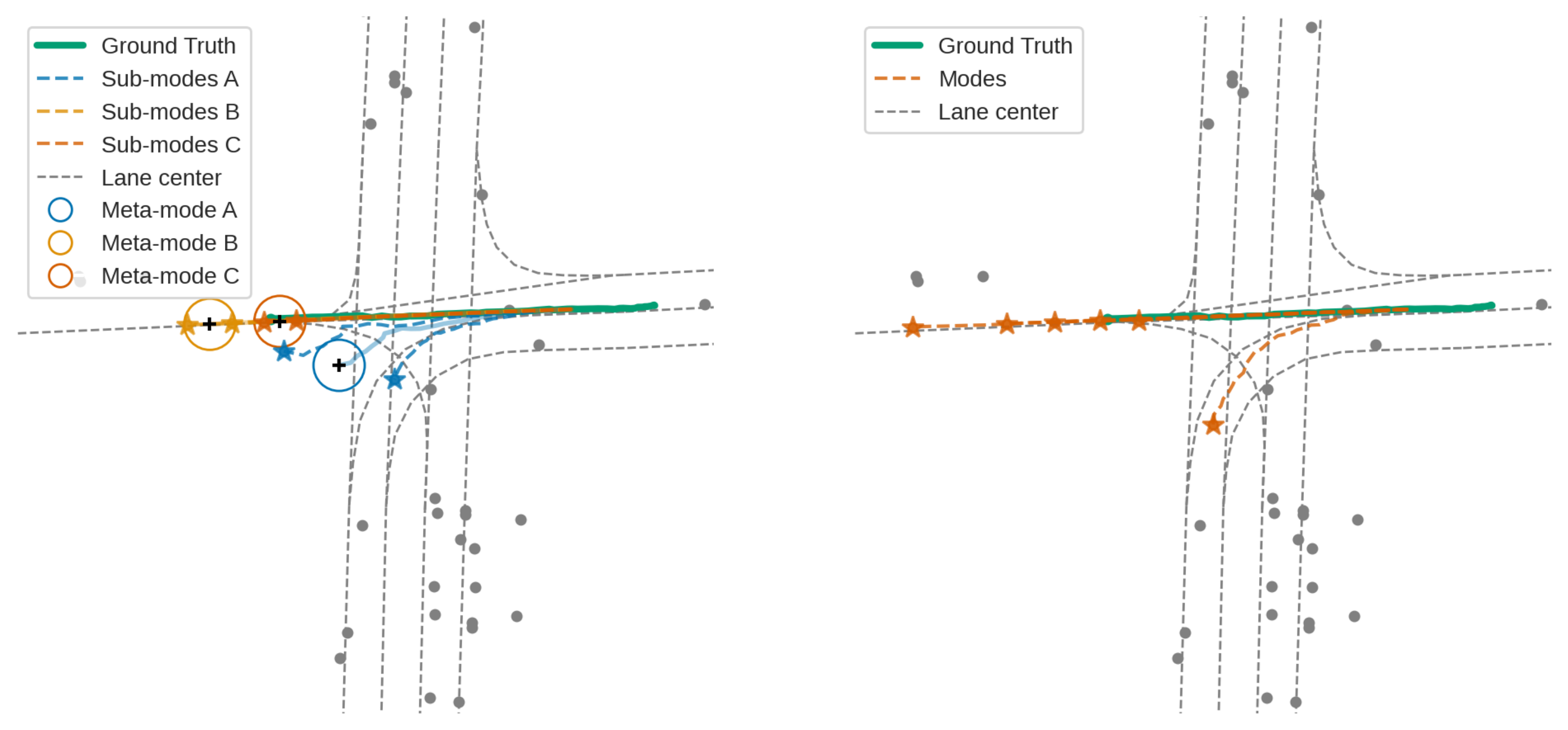} 
        \caption{Another intersection scenario}
        \label{fig:quali_4}
    \end{subfigure}
    \caption{\textbf{Qualitative results with AutoBots backbone on Argoverse 1.} We compare an AutoBots model trained with its original loss ($K=6$) compared with our HWTA loss ($\gamma=0.8$) with $3$ \textit{meta-modes} (i.e., $K^{\star}=3$ and $K'=2$).}
    \label{ann:qualitative}
\end{figure*}

\cref{ann:qualitative} provides additional trajectory forecasting examples on the Argoverse 1 dataset. We display predictions from an AutoBots trained using our new loss HWTA and a classical AutoBots model.
The first thing we observe is that the predictions from our model are less scattered than the other. Indeed, they are closer to each other and the ground truth, which explains why we reached higher performance on $\text{mADE}_1$, $\text{mFDE}_1$ and the $\text{NLL}_k$ metrics. We also note that the \textit{sub-modes} belonging to the same \textit{meta-modes} are near each other, as announced in the paper. With this knowledge, we can expect the average of several \textit{sub-modes} (i.e., a \textit{meta-mode}) to be a more robust prediction, as it might be in higher-density areas.
Finally, we show a failure example, where due to the bad position of one \textit{sub-mode}, the \textit{meta-mode} A is misplaced. We argue that this issue could occur for likely trajectory candidates and that one could easily compute the intra-mode distances to see whether the cluster is coherent. Moreover, adding more \textit{sub-modes} per \textit{meta-mode} should alleviate this issue as one bad \textit{sub-mode} will have less effect.

\end{document}